\pdfoutput=1

\documentclass[11pt]{article}

\usepackage[final]{coling}
\usepackage{multirow}
\usepackage{amssymb}
\usepackage{amsmath}
\usepackage{graphicx}
\usepackage{subfigure}
\usepackage{wrapfig}
\usepackage{float}
\usepackage{times}
\usepackage{latexsym}

\usepackage[T1]{fontenc}

\usepackage[utf8]{inputenc}

\usepackage{microtype}

\usepackage{inconsolata}

\usepackage{graphicx}

\title{Distance-Adaptive Quaternion Knowledge Graph Embedding with Bidirectional Rotation}

\author{
 \textbf{Weihua Wang\textsuperscript{1,2,3,}\thanks{Corresponding Author. Email: \href{mailto:wangwh@imu.edu.cn}{wangwh@imu.edu.cn.}}},
 \textbf{Qiuyu Liang\textsuperscript{1}},
 \textbf{Feilong Bao\textsuperscript{1,2,3}},
 \textbf{Guanglai Gao\textsuperscript{1,2,3}}
\\
\\
 \textsuperscript{1}College of Computer Science, Inner Mongolia University, Hohhot, China
 \\
 \textsuperscript{2}National and Local Joint Engineering Research Center of Intelligent 
 Information 
 \\
 Processing Technology for Mongolian, Hohhot, China
 \\
 \textsuperscript{3}Inner Mongolia Key Laboratory of Multilingual Artificial Intelligence Technology, Hohhot, China
\\
}

\begin{document}
\maketitle
\begin{abstract}

Quaternion contains one real part and three imaginary parts, which provided a more expressive hypercomplex space for learning knowledge graph.
Existing quaternion embedding models measure the plausibility of a triplet through either semantic matching or geometric distance scoring functions.
However, it appears that semantic matching diminishes the separability of entities, while the distance scoring function weakens the semantics of entities.
To address this issue, we propose a novel quaternion knowledge graph embedding model. 
Our model combines semantic matching with the geometric distance of entities to better measure the plausibility of triplets.
Specifically, in the quaternion space, we perform a right rotation on head entity and a reverse rotation on tail entity to learn rich semantic features.
We then utilize distance-adaptive translations to learn geometric distance between entities.
Furthermore, we provide mathematical proofs to demonstrate our model can handle complex logical relationships.
Extensive experimental results and analyses show our model significantly outperforms previous models on well-known knowledge graph completion benchmark datasets.
Our code is available at \url{https://github.com/llqy123/DaBR}.
\end{abstract}\

\section{Introduction}
\label{intro}
Knowledge graphs (KGs) \cite{10577554} are powerful tools for representing valid factual triplets by capturing entities and their relationships in a graphical format. 
Owing to the well-structured of graphs, KGs are often used for various Natural Language Processing tasks, such as question answering \cite{mendes-etal-2024-application,faldu-etal-2024-retinaqa}, entity alignment \cite{wang2024unifyingdualspaceembeddingentity, wang2024gsea}, KG-based recommendation \cite{liang12024knowledge} and KG enhanced Large Language Model \cite{wen-etal-2024-mindmap}.

However, KGs are usually incomplete and the incompleteness limits their application.
As an effective tool for predicting missing facts, knowledge graph completion (KGC) has received considerable attention from researchers.
Typically, researchers transform KGC tasks into knowledge graph embeddings (KGEs).
KGE refers to learning representations of entities and relations in a low-dimensional space while preserving the graph's inherent structure and semantic properties.
In this representation space, a scoring function can be defined to measure the plausibility of each triplet, where valid triplets should receive higher scores than these invalid ones.

\begin{figure}
\centering 
\subfigure[QuatE]{
\includegraphics[width=0.485\linewidth]{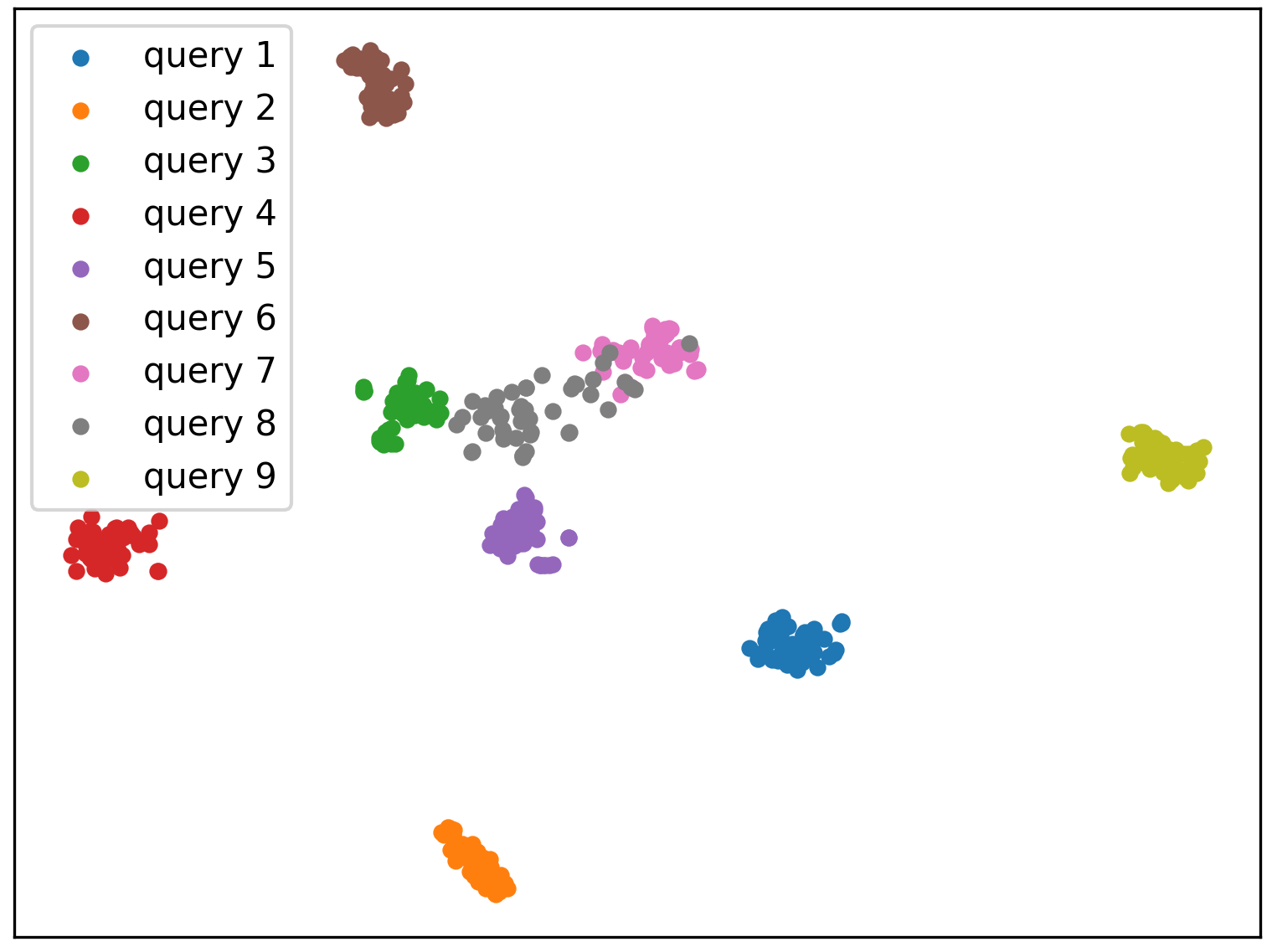}}\subfigure[TransERR]{
\includegraphics[width=0.485\linewidth]{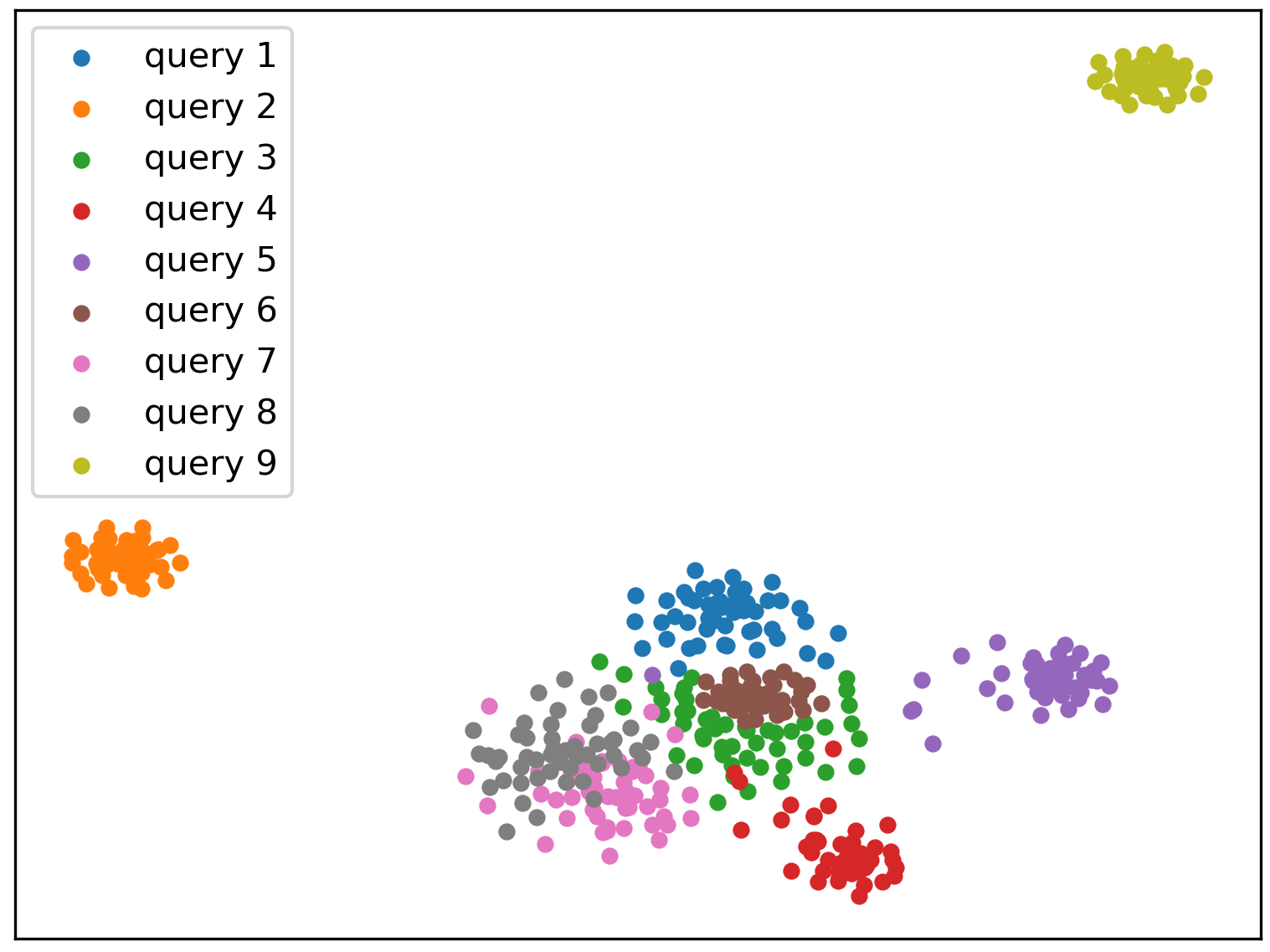}}
\caption{The visualization embedding of QuatE and TransERR models after 100 epochs training.
Points in the same color represent tail entities that have the same $(h_r, r_j )$ (query) context.
}
\label{Fig:headfigure}
\end{figure}

Quaternion contains one real part and three imaginary parts, which providing a more expressive space for learning embeddings of entities and relations.
Rotation in the quaternion space is often used to model the KGs.
For example, QuatE \cite{zhang2019quaternion} learns semantic information about entities by treating relations as rotations from head entities to tail entities.
TransERR \cite{li-etal-2024-transerr-translation} encodes the KG by rotating the head and tail entities with their corresponding unit quaternions.
These models use either semantic matching or distance scoring functions to measure the plausibility of the triplet, respectively.
However, it appears that semantic matching diminishes the separability of entities, while the distance scoring function weakens the semantics of entities.
For example, we visualized the results for the same query in Figure \ref{Fig:headfigure} \footnote{For more information about queries, see Section \ref{sec:vis}.}.
Specifically, as shown in Figure \ref{Fig:headfigure}, we observe that QuatE model overlaps some queries when using semantic matching as a scoring function.
The entities of TransERR using the distance scoring function are also indistinguishable from each query.

To address this issue, we propose a \textbf{D}istance-\textbf{a}daptive quaternion knowledge graph embedding with \textbf{B}idirectional \textbf{R}otation model, named as \textbf{DaBR}. 
Our model combines semantic matching with the geometric distance of entities to better measure the plausibility of triplets.
Specifically, in the quaternion space, we perform a right rotation on the head entity and a reverse rotation on the tail entity to learn rich semantic features.
This process is called bidirectional rotation.
We conducted extensive experiments on multiple well-known benchmark datasets for knowledge graph completion task.
The experimental results and analyses demonstrated the effectiveness and robustness of our model.

Our contributions are summarized as follows:
\begin{itemize}
    \item We propose performing a right rotation on the head entity and a reverse rotation on the tail entity to learn rich semantic features.
    \item We propose learning the embedding distance between entities by incorporating distance adaptive translations.
    \item We provide mathematical proofs to demonstrate that our model can handle rich logical relationships.
    \item Extensive experiments show that our model provides consistent and significant improvements over previous models in most metrics.
\end{itemize}

\section{Related Work}

For KGE models, the design of the scoring function directly affects these models' performance and effectiveness. 
Based on the calculation methods of scoring functions in previous models, KGE scoring functions can mainly be categorized into semantic matching- and geometric distance-based.

\textbf{Semantic matching.} 
Semantic matching scoring functions capture the interactions between entities and relations through inner products on embedding vectors. 
The hypothesis is that entities connected by relations are close to each other in the semantic space. 
For example, QuatE \cite{zhang2019quaternion} obtains semantic information about entities through the Hamiltonian rotation of the head entity on the relation in quaternion space. 
DualE \cite{Cao_Xu_Yang_Cao_Huang_2021} further enhances QuatE to model knowledge graphs in dual quaternion space. 
QuatRE \cite{10.1145/3487553.3524251} associates each relation with two relation-aware rotations, which are used to rotate the quaternion embeddings of the head and tail entities, respectively.
ConvQE \cite{liang2024effective} investigates the potential of quaternion convolution in knowledge graph embedding.

A common feature of these models is the computation of the inner product between the head entity and the tail entity after a relation transformation. 
However, these models overlook the geometric distance properties between entities in the knowledge graph, which leads to distorted embeddings of the learned entities.

\textbf{Geometric distance.} 
Geometric distance scoring functions assess the plausibility of triplets by calculating the distances between embedding vectors in the representation space.
The goal of this scoring function is to keep the head/tail entity vector closer to the tail/head entity vector after being transformed through the relation vector.
For example, TransE \cite{bordes2013translating}, considered the first model to employ a geometric distance scoring function, assumes that triplets $(h,r,t)$ in knowledge graphs should satisfy the expression $h+r\approx t$.
However, TransE struggles with more complex relation types, such as one-to-many (1-to-N), many-to-one (N-to-1) and many-to-many (N-to-N).

To address this limitation, several models using distance-based scoring functions have been proposed. 
For example, Rotate3D \cite{10.1145/3340531.3411889} maps entities to a 3D space, defining the relation as a rotation from the head entity to the tail entity. 
Trans4E \cite{NAYYERI2021530} performs rotations and translations in a quaternion space. 
RotateCT \cite{dong-etal-2022-rotatect} transforms entity coordinates and represents each relation as a rotation in complex space. 
Rotate4D \cite{LE2023110400} employs two distinct rotational transformations to align the head embedding with the tail embedding. 
DCNE \cite{dong-etal-2024-dual-complex} maps entities to the dual complex number space, using rotations in the 2D space through the multiplication of dual complex numbers to represent relations. 
TransERR \cite{li-etal-2024-transerr-translation} encodes knowledge graphs by rotating the head and tail entities with their corresponding unit quaternions.

A common feature of these models is that the plausibility of the triplets is evaluated by calculating the distance between the head entity and the tail entity after transformation. 
However, these models do not consider information about entities within the semantic space, leading to performance degradation.

\section{Preliminaries}

This section begins with a definition of the knowledge graph completion task, followed by a brief background on quaternion algebra.

\subsection{Knowledge Graph Completion}

Knowledge graph completion is the task of predicting missing elements in a triplet $(h,r,t)$. 
This task can be broken down into three sub-tasks: predicting the head entity $(?,r,t)$, predicting the relation $(h,?,t)$, and predicting the tail entity $(h,r,?)$. 
Following previous research, our work focuses on predicting the head $(?,r,t)$ and tail $(h,r,?)$ entities.
It is because relation information is needed in the training process.

\subsection{Quaternion Algebra}
The quaternion extends the complex number system to four dimensions.
In $n$-dimensional quaternion space $\mathbb{Q}^n$, a quaternion $\mathbf{p} \in \mathbb{Q}^n$ consists of one real component and three imaginary components.
It can be formalized as: $\mathbf{p} = a + b\mathbf{i} + c\mathbf{j} + d\mathbf{k}$, where $a, b, c, d \in \mathbb{R}^\frac{n}{4}$ are real numbers and $\mathbf{i, j, k }$ are imaginary units.
The imaginary part satisfies the Hamilton’s rules \cite{hamilton1844theory}: $\mathbf{i}^2=\mathbf{j}^2=\mathbf{k}^2=\mathbf{ijk}=-1$.

\textbf{Addition.}
Given two quaternions $\mathbf{p} = a + b\mathbf{i} + c\mathbf{j} + d\mathbf{k}$ and $\mathbf{q} = e + f\mathbf{i} + g\mathbf{j} + h\mathbf{k} \in\mathbb{Q}^n$, quaternion addition is defined as:
\begin{equation}
\label{equ:add}
    \mathbf{p} + \mathbf{q} = (a+e)+(b+f)\mathbf{i}+(c+g)\mathbf{j}+(d+h)\mathbf{k}
\end{equation}

\textbf{Norm.} 
The normalization of quaternions $\parallel \mathbf{p} \parallel \in \mathbb{Q}^n$ can be defined by the following:
\begin{equation}
\label{equ:norm}
    \parallel \mathbf{p} \parallel =\sqrt{a^2+b^2+c^2+d^2}.
\end{equation}

\textbf{Inverse.} 
The inverse of quaternions $\parallel \mathbf{p} \parallel \in \mathbb{Q}^n$ can be defined by the following:
\begin{equation}
\begin{aligned}
\label{equ:inverse}
    \mathbf{p}^{-1}=\frac{\mathbf{\bar{p}}}{\parallel \mathbf{p} \parallel ^2},\mathbf{\bar{p}} \;=a - b\mathbf{i} - c\mathbf{j} - d\mathbf{k},
\end{aligned}
\end{equation}
where $\mathbf{\bar{p}}\in\mathbb{Q}^n$ is the conjugate of $\mathbf{p}\in\mathbb{Q}^n$.

\textbf{Hamilton product.} 
Given two quaternions $\mathbf{p}$ and $\mathbf{q}$. 
The quaternion rotation of these two quaternions can be performed by the Hamilton product:
\begin{equation}
\label{equ:ham}
\begin{aligned}
    \mathbf{p}\otimes \mathbf{q} = &(a \circ e - b \circ f - c \circ g - d \circ h) +\\
                 &(b \circ e + a \circ f + c \circ h - d \circ g)\mathbf{i} + \\
                 &(c \circ e + a \circ g + d \circ f - b \circ h)\mathbf{j}+\\
                 &(d \circ e + a \circ h + b \circ g - c \circ f)\mathbf{k},
\end{aligned}
\end{equation}
where $\circ$ denotes the element-wise product.


\section{Methodology}

\begin{figure*}
\centering 
\subfigure[QuatE]{
\label{Fig:QuatE}
\includegraphics[width=0.32\linewidth]{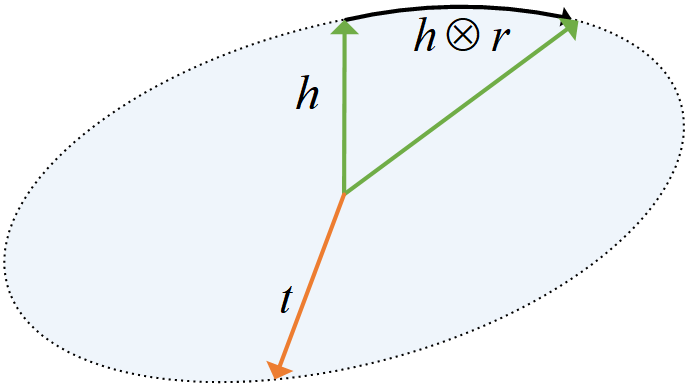}}\subfigure[QuatRE]{
\label{Fig:QuatRE}
\includegraphics[width=0.32\linewidth]{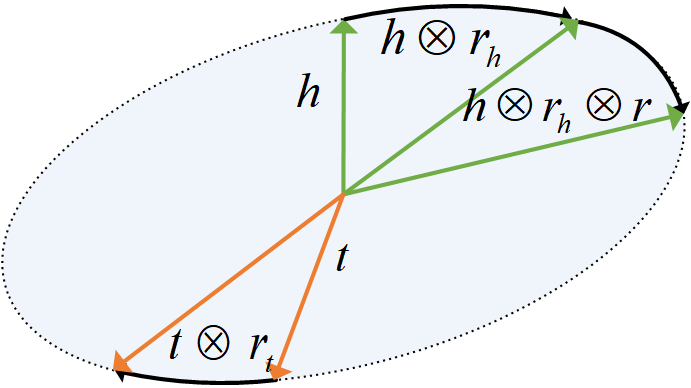}}\subfigure[DaBR (ours)]{
\label{Fig:DaBR}
\includegraphics[width=0.32\linewidth]{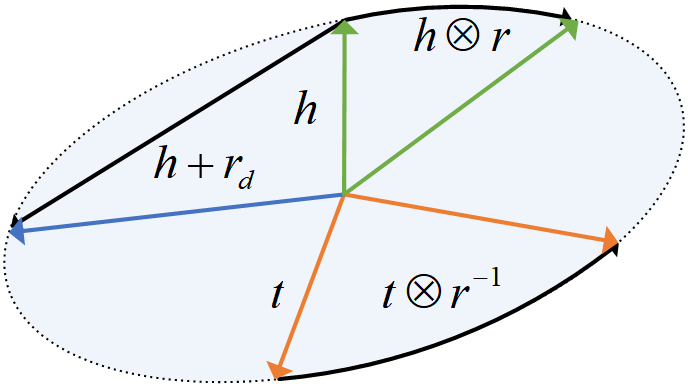}}
\caption{The comparison of modeling entity semantics of QuatE, QuatRE and DaBR.
These models learn the embeddings of knowledge graphs in quaternion spaces.
$\otimes$ denotes the Hamilton product (Equation \ref{equ:ham}).
}
\label{Fig:compare}
\end{figure*}

In this section, we describe our model in detail, which consists of two main parts: 
\begin{itemize}
    \item \textbf{Bidirectional rotation}: Performing a right rotation on the head entity and a reverse rotation on the tail entity to learn the rich semantic features.
    \item \textbf{Distance-adaptation}: Incorporating a distance adaptive translation to learn the geometric distance between entity embeddings.
\end{itemize}

\subsection{Symbol Description}
A knowledge graph $\mathcal{G}=:\{(h,r,t)\} \in \mathcal{E} \times \mathcal{R} \times \mathcal{E}$ is a collection of triplet, where $\mathcal{E}$ and $\mathcal{R}$ are the entity set and relation set.
$|\mathcal{E}| $ and $ |\mathcal{R}| $ represent the number of entities and relations, respectively.
Given a triplet $(h,r,t)$, the embeddings of head entity $\mathbf{h}$, relation $\mathbf{r}$ and tail entity $\mathbf{t}$ can be represented by quaternions:
\begin{equation}
\begin{aligned}
    \mathbf{h} &= a_h + b_h \mathbf{i} + c_h \mathbf{j} + d_h \mathbf{k} \\
    \mathbf{r} &= p + q \mathbf{i} + u \mathbf{j} + v \mathbf{k} \\
    \mathbf{t} &= a_t + b_t \mathbf{i} + c_t \mathbf{j} + d_t \mathbf{k} \\
\end{aligned}
\end{equation}

\subsection{Part One: Bidirectional Rotation}
In Figure \ref{Fig:compare}, we show the differences between our proposed bidirectional rotation and previous methods when modeling entity semantics.
Specifically, QuatE (Figure \ref{Fig:QuatE}) performs a right rotation for head entity.
QuatRE (Figure \ref{Fig:QuatRE}) performs two times right rotation for head entity and a right rotation for tail entity.
Our model (Figure \ref{Fig:DaBR}) performs a right rotation for head entity and a reverse rotation for tail entity.

We first normalize the relation quaternion $\mathbf{r}$ to a unit quaternion $\mathbf{r}^\lhd $ to eliminate the scaling effect by dividing by its norm (Equation \ref{equ:norm}):
\begin{equation}
    \mathbf{r}^\lhd = \frac{\mathbf{r}}{\parallel \mathbf{r} \parallel }=\frac{p + q \mathbf{i} + u \mathbf{j} + v \mathbf{k}}{\sqrt{p^2 + q^2 + u^2 + v^2}}.
\end{equation}

Then, the head entity $\mathbf{h}$ is right rotated using the relation $\mathbf{r}^\lhd$, i.e., the entity vector and the relation vector do a Hamilton product (Equation \ref{equ:ham}):
\begin{equation}
    \mathbf{h}^{\prime} = \mathbf{h} \otimes \mathbf{r}^{\lhd}. 
\end{equation}

Similarly, the inverse of the relation unit quaternion $\mathbf{r}^\lhd$ is used to make a reverse rotation of the tail entity $\mathbf{t}$:
\begin{equation}
    \mathbf{t}^{\prime} = \mathbf{t} \otimes \mathbf{r}^{\lhd -1}. 
\end{equation}

Since $\mathbf{r}^ \lhd$ is a unit quaternion, we have:
\begin{equation}
\begin{aligned}
    \mathbf{t}^{\prime} = \mathbf{t} \otimes \mathbf{r}^{\lhd -1} = \mathbf{t} \otimes \mathbf{\bar{r}^\lhd},
\end{aligned}
\end{equation}
where $\mathbf{\bar{r}}^\lhd$ is the conjugate of $\mathbf{r}^\lhd$.

Therefore, the scoring function $s(h,r,t)$ for the bidirectional rotation modeling entity semantics is defined by:
\begin{equation}
    s(h,r,t) = \mathbf{h}^{\prime} \circ \mathbf{t}^{\prime}=\mathbf{h} \otimes \mathbf{r}^{\lhd} \circ \mathbf{t} \otimes \mathbf{\bar{r}^\lhd},
\end{equation}

\subsection{Part Two: Distance-Adaptation}
As shown in Figure \ref{Fig:compare}, the previous QuatE (Figure \ref{Fig:QuatE}) and QuatRE (Figure \ref{Fig:QuatRE}) can only learn the semantic information of an entity but ignore the geometric distance attribute of an entity.
Our DaBR effectively addresses this limitation by adding a distance-adaptation (Figure \ref{Fig:DaBR}).

Therefore, to model the geometric distance information, we initialize a distance-adaptive relation embedding $\mathbf{r_d} = p_d + q_d \mathbf{i} + u_d \mathbf{j} + v_d \mathbf{k}$.
Finally, the geometric distance part scoring function $d(h,r,t)$ is defined as:
\begin{equation}
    d(h,r,t) = \parallel\mathbf{h} + \mathbf{r_d} - \mathbf{t} \parallel_1,
\end{equation}
where $\parallel \cdot \parallel_1$ represents the $\ell_1$ norm. 
Despite its simplicity, we find that the proposed method is effective enough in providing distance information for our model.

\subsection{Scoring Function}
After obtaining the scoring functions for modeling entity semantics and entity geometric distances, respectively.
We fuse these scoring functions into a new scoring function for model training:
\begin{equation}
\begin{aligned}
    &\phi(h,r,t) = s(h,r,t) + \lambda d(h,r,t) \\
    &= \mathbf{h} \otimes \mathbf{r}^{\lhd} \cdot \mathbf{t} \otimes \mathbf{\bar{r}^\lhd} + \lambda \parallel\mathbf{h} + \mathbf{r_d} - \mathbf{t} \parallel_1,
\end{aligned} 
\end{equation}
where $s(h,r,t)$ represents the semantic matching scoring function, $d(h,r,t)$ represents the geometric distance scoring function, and $\lambda \in \mathbb{R}$ is an adaptive parameter that learned by our model.

\subsection{Loss Function}
Following \citet{10.5555/3045390.3045609}, we formulate the task as a classification problem, and the model parameters are learned by minimizing the following regularized logistic loss:
\begin{equation}
\begin{aligned}
    \mathcal{L}=&\displaystyle\sum_{r(h,t)\in \Omega \cup \Omega ^-}^{}\text{log}(1+\text{exp}(-Y_{hrt}\phi (h,r,t)))\\
    &+\eta_1\parallel \mathbf{E}\parallel^2_2 +\eta_2\parallel \mathbf{R}\parallel^2_2,
\end{aligned}
\end{equation}
where $\mathbf{E}$ and $\mathbf{R}$ denote the embedding of all entities and relations.
Here we use the $\ell_2$ norm with regularization rates $\eta_1$ and $\eta_2$ to regularize $E$ and $R$, respectively.
$\Omega ^-$ is sampled from the unobserved set $\Omega ^\prime$ using uniform sampling.
$Y_{hrt} \in \{-1,1\}$ represents the corresponding label of the triplet $(h,r,t)$.

\begin{table*}
\centering
\resizebox{\textwidth}{!}{
\begin{tabular}{clcccccccccc}
\hline
\multirow{2}{*}{\textbf{SF}} &\multirow{2}{*}{\textbf{Model}}      & \multicolumn{5}{c}{WN18RR}   & \multicolumn{5}{c}{FB15k-237}  \\ 
&  & MR($\downarrow $)   & MRR     & H@10  & H@3 & \multicolumn{1}{c}{H@1}  & MR($\downarrow $)    & MRR     & H@10  & H@3 & \multicolumn{1}{c}{H@1}         \\ 
\hline
\multirow{4}{*}{\textbf{SM}}
& TuckER \citeyearpar{balazevic-etal-2019-tucker} &- &.470 &.526 &.482 &\multicolumn{1}{c|}{.443} &- &.358 &.544 &.394 &.266 \\
&QuatE \citeyearpar{zhang2019quaternion}	&2314	&.488	&.582	&.508	&\multicolumn{1}{c|}{.438}		&\underline{87}	&.348	&.550	&.382	&.248 \\
&DualE \citeyearpar{Cao_Xu_Yang_Cao_Huang_2021}	&2270	&.492	&.584	&.513	&\multicolumn{1}{c|}{.444}		&91	&.365	&.559	&.400	&.268\\
&QuatRE \citeyearpar{10.1145/3487553.3524251}	&1986	&.493	&.592	&.519	&\multicolumn{1}{c|}{.439}		&88	&\underline{.367}	&\underline{.563}	&\underline{.404}	&.269\\
&ConvQE \citeyearpar{liang2024effective} &- &.487	&.563	&.502	& \multicolumn{1}{c|}{.447} 	&- & .366 &.551	&.402	&\underline{.273}	\\
\hline
\multirow{10}{*}{\textbf{GD}}
&ATTH \citeyearpar{chami-etal-2020-low} &- &.486 &.573 &.499 &\multicolumn{1}{c|}{.443}  &- &.348 &.540 &.384 &.252 \\
&Rotate3D \citeyearpar{10.1145/3340531.3411889}	&3328	&.489	&.579	&.505	&\multicolumn{1}{c|}{.442}		&165	&.347	&.250	&.543	&.385\\
&Trans4E \citeyearpar{NAYYERI2021530} &1755	&.469	&.577	&.487	&\multicolumn{1}{c|}{.416}		&158	&.332	&.527	&.366	&.236\\
&RotateCT \citeyearpar{dong-etal-2022-rotatect}	&3285	&.492	&.579	&.507	&\multicolumn{1}{c|}{.448}		&171	&.347	&.537	&.382	&.251\\
&Rotate4D \citeyearpar{LE2023110400}	&3167	&.499	&.587	&.518	&\multicolumn{1}{c|}{\textbf{.455}}		&181	&.353	&.547	&.391	&.257\\
&CompoundE \citeyearpar{ge-etal-2023-compounding}	&-	&.491	&.576	&.508	&\multicolumn{1}{c|}{\underline{.450}}		&-	&.357	&.545	&.393	&.264\\
&HAQE \citeyearpar{10650007} &- &.496  & .584 & .512	&\multicolumn{1}{c|}{\underline{.450}}		&- &.343 &.535	&.379	&.247 \\
&DCNE \citeyearpar{dong-etal-2024-dual-complex}	&3244	&.492	&.581	&.510 	&\multicolumn{1}{c|}{.448}		    &169	&.354	&.547	&.393	&.257\\
&FHRE \citeyearpar{Liang2024FullyHR} &- &.494 &.563 &.510	&\underline{.450} &- &.345	&.528	&.375	&.255  \\		 
&TransERR \citeyearpar{li-etal-2024-transerr-translation}	&\underline{1167}	&\underline{.501}	&\underline{.605}	&\underline{.520} 	&\multicolumn{1}{c|}{\underline{.450}}		&125	&.360 	&.555	&.396	&.264\\
\hline
\multirow{1}{*}{\textbf{SG}}							
&\textbf{DaBR (ours)} &\textbf{899}	&\textbf{.510}	&\textbf{.622}	&\textbf{.538}	&\multicolumn{1}{c|}{\underline{.450}} & \textbf{83} &\textbf{.373}	&\textbf{.572}	&\textbf{.410}	&\textbf{.274}\\
\hline
\end{tabular}
}
\caption{\label{tab:mainresult}
Knowledge graph completion results on WN18RR and FB15k-237 datasets.
Best results are in bold and second best results are underlined.
\textbf{SF} indicates the scoring function, \textbf{SM} indicates semantic matching scoring function, \textbf{GD} indicates geometric distance scoring function, and \textbf{SG} indicates our semantic matching and geometric distance scoring function.
``-'' indicates that there is no result reported.
The same settings apply to Table \ref{tab:subresult}.
}
\end{table*}

\subsection{Discussion}
As described in \citet{chami-etal-2020-low}, there are complex logical relationships (such as symmetry, antisymmetry, inversion and composition relationships) in the knowledge graph.
In this part, we analyze the ability of our DaBR to infer these relationships.

\noindent 
\textbf{Lemma 1} \textit{DaBR can infer the symmetry relationship pattern. (See proof in Appendix \ref{pr1})}

\noindent 
\textbf{Lemma 2} \textit{DaBR can infer the antisymmetry relationship pattern. (See proof in Appendix \ref{pr2})}

\noindent 
\textbf{Lemma 3} \textit{DaBR can infer the inversion relationship pattern. (See proof in Appendix \ref{pr3})}

\noindent 
\textbf{Lemma 4} \textit{DaBR can infer the composition relationship pattern. (See proof in Appendix \ref{pr4})}

\section{Experiments}
In this section, we first introduce the datasets, evaluation protocol, implementation details and baselines. 
Subsequently, we evaluate our model on four benchmark datasets.

\noindent
\textbf{Datasets}.
To verify the effectiveness and robustness of our model, we conducted extensive experiments on four standard knowledge graph completion datasets including WN18RR \cite{Dettmersconve}, FB15k-237 \cite{toutanova-chen-2015-observed}, WN18 \cite{bordes2013translating} and FB15k \cite{bordes2013translating}.
The WN18 and FB15k datasets are known to suffer from a data leakage problem, which causes models to easily inferred and consequently performing well on metrics.
WN18RR and FB15k-237 were derived as subsets of WN18 and FB15k respectively. 
These datasets are designed to address data leakage concerns and thereby present a more realistic prediction task.
The detailed statistics of the four standard datasets are shown in Appendix \ref{app:datasta}.

\noindent
\textbf{Evaluation protocol}.
Similar to previous work \cite{zhang2019quaternion, li-etal-2024-transerr-translation}, we employed the filtered evaluation setup described in reference \cite{bordes2013translating} to filter out real triplets during the evaluation process. 
This was done to avoid flawed evaluations.
We used evaluation metrics encompassed Mean Rank (MR), Mean Reciprocity Rating (MRR) and Hits@n (n=1, 3 or 10).
Where a smaller value on the MR indicates a better model.
The final scoring model on the test set is derived from the model with the highest Hits@10 score on the validation set.

\noindent
\textbf{Implementation details}.
We conduct all our experiments on a single NVIDIA GeForce RTX 4090 with 24GB of memory. 
The ranges of the hyper-parameters for the grid search are set as follows: the embedding dimension ($dim$) is selected from \{300, 400, 500\}; the learning rate ($lr$) is chosen from \{0.01, 0.02, 0.05, 0.1\}; and the number of negative triplets sampled ($neg$) per training triplet is selected from \{5, 10\}. 
The regularization rates $\eta_1$ and $\eta_2$ are adjusted within \{0.01, 0.05, 0.1, 0.5\}. 
We create 100 batches of training samples for different datasets. 
We optimize the loss function by utilizing Adagrad \cite{10.5555/1953048.2021068}.
All our hyper-parameters are provided in Appendix \ref{app:optimal}.

It is worth noting that our models \textbf{do not} employ the training strategies of self-adversarial negative sampling \cite{sun2018rotate} or N3 regularization with reciprocal learning \cite{pmlr-v80-lacroix18a}.

\begin{table*}[ht]
\renewcommand\arraystretch{1.1}
\centering
\small
\begin{tabular}{clcccccccccc}
\hline
\multirow{2}{*}{\textbf{SF}} &\multirow{2}{*}{\textbf{Model}}      & \multicolumn{5}{c}{WN18}   & \multicolumn{5}{c}{FB15k}  \\ 
&  & MR($\downarrow $)   & MRR     & H@10  & H@3 & \multicolumn{1}{c}{H@1}  & MR($\downarrow $)    & MRR     & H@10  & H@3 & \multicolumn{1}{c}{H@1}         \\ 
\hline
\multirow{4}{*}{\textbf{SM}}
& TuckER \citeyearpar{balazevic-etal-2019-tucker} &- &\underline{.953} &.958 &.955 &\multicolumn{1}{c|}{\textbf{.949}} &- &.795 &.892 &.833 &.741 \\
&QuatE \citeyearpar{zhang2019quaternion}	&162	&.950	&.959	&.954	&\multicolumn{1}{c|}{.945}		&\textbf{17}	&.782	&\textbf{.900}	&.835	&.711 \\
&DualE \citeyearpar{Cao_Xu_Yang_Cao_Huang_2021}	&156	&.952	&.962	&.956	&\multicolumn{1}{c|}{\underline{.946}}		&21	&.813	&\underline{.896}	&\underline{.850}	&.766\\
&QuatRE \citeyearpar{10.1145/3487553.3524251}	&116	&.939	&.963	&.953	&\multicolumn{1}{c|}{\underline{.946}}		&21	&.808	&\underline{.896}	&.851	&.751\\

\hline
\multirow{6}{*}{\textbf{GD}}
&Rotate3D \citeyearpar{10.1145/3340531.3411889}	&214	&.951	&.961	&.953	&\multicolumn{1}{c|}{.945}		&39	&.789	&.887	&.832	&.728 \\
&Trans4E \citeyearpar{NAYYERI2021530} &175	&.950	&.960	&.953	&\multicolumn{1}{c|}{.944}		&47	&.767	&.892	&.834	&.681
\\
&RotateCT \citeyearpar{dong-etal-2022-rotatect}	&201	&.951	&.963	&.956	&\multicolumn{1}{c|}{.944}		&34	&.794	&.888	&.834	&.737
\\
&Rotate4D \citeyearpar{LE2023110400}	&173	&.952	&.963	&.956	&\multicolumn{1}{c|}{\underline{.946}}		&37	&.790 	&.887	&.831	&.732
\\
&DCNE \citeyearpar{dong-etal-2024-dual-complex}	&192	&.952	&.963	&.955	&\multicolumn{1}{c|}{.945}		&34	&.798	&.888	&.835	&.745
\\
&TransERR \citeyearpar{li-etal-2024-transerr-translation}	&\underline{82}	&\underline{.953}	&\underline{.965}	&\underline{.957}	&\multicolumn{1}{c|}{.945}		&41	&\underline{.815}	&\underline{.896}	&.848	&\underline{.767} \\
\hline
\multirow{1}{*}{\textbf{SG}}							
&\textbf{DaBR (ours)} &\textbf{56}	&\textbf{.954}	&\textbf{.966}	&\textbf{.959}	&\multicolumn{1}{c|}{\underline{.946}} & \underline{18} &\textbf{.819}	&\textbf{.900}	&\textbf{.854}	&\textbf{.769}\\
\hline
\end{tabular}
\caption{\label{tab:subresult}
Knowledge graph completion results on WN18 and FB15k datasets.}
\end{table*}

\begin{table*}[]
\renewcommand\arraystretch{1.05}
\setlength{\tabcolsep}{2.5pt}
\centering
\resizebox{\textwidth}{!}{
\begin{tabular}{lcccccccccccccccc}
\hline
\multirow{2}{*}{\textbf{Model}}    & \multicolumn{4}{c}{WN18RR}   & \multicolumn{4}{c}{FB15k-237} & \multicolumn{4}{c}{WN18}   & \multicolumn{4}{c}{FB15k}   \\

& MRR     & H@10  & H@3 & \multicolumn{1}{c}{H@1}  & MRR     & H@10  & H@3 & \multicolumn{1}{c}{H@1} & MRR     & H@10  & H@3 & \multicolumn{1}{c}{H@1} & MRR     & H@10  & H@3 & \multicolumn{1}{c}{H@1}         \\ 
\hline

DaBR &\textbf{.510}	&\textbf{.622}	&\textbf{.538}	&\multicolumn{1}{c|}{\textbf{.450}}  &\textbf{.373}	&\textbf{.572}	&\textbf{.410}	&\multicolumn{1}{c|}{\textbf{.274}} &\textbf{.954}	&\textbf{.966}	&\textbf{.959}	&\multicolumn{1}{c|}{\textbf{.946}}  &\textbf{.819}	&\textbf{.900}	&\textbf{.854}	&\textbf{.769}\\
\textit{Variant I} &.505  &.617 &.532 &\multicolumn{1}{c|}{.445} &.370 &.569 &.404 &\multicolumn{1}{c|}{.272} &.953 	 &.964 	 &.956      &\multicolumn{1}{c|}{.943}  &.816  	 &.894 	 &.844   &.766 \\
\textit{Variant II} &.495  &.580 &.512 &\multicolumn{1}{c|}{.445} &.368 &.566 &.402 &\multicolumn{1}{c|}{.270} &.947  &.960 &.954 &\multicolumn{1}{c|}{.937}  &.801  	 &.890 	 &.847   &.751\\
\hline
\end{tabular}
}
\caption{\label{tab:ablationresult}
Ablation results for all datasets.}
\end{table*}

\noindent
\textbf{Baselines}.
To verify the effectiveness of our model, we compared DaBR with several powerful baseline models, including both well-known and recently proposed ones with outstanding results.
We divide these models according to the scoring function: 

\textbf{1) Semantic Matching:} TuckER \cite{balazevic-etal-2019-tucker}, QuatE \cite{zhang2019quaternion}, DualE \cite{Cao_Xu_Yang_Cao_Huang_2021}, QuatRE \cite{10.1145/3487553.3524251} and ConvQE \cite{liang2024effective}.

\textbf{2) Geometric Distance:} ATTH \cite{chami-etal-2020-low}, Rotate3D \cite{10.1145/3340531.3411889}, Trans4E \cite{NAYYERI2021530}, RotateCT \cite{dong-etal-2022-rotatect}, Rotate4D \cite{LE2023110400}, CompoundE \cite{ge-etal-2023-compounding}, HAQE \cite{10650007}, DCNE \cite{dong-etal-2024-dual-complex}, FHRE \cite{Liang2024FullyHR} and TransERR \cite{li-etal-2024-transerr-translation}. 

For a fair comparison, we report the optimal results for these baselines from the original papers.

\subsection{Main Results}
The main results of our DaBR and the baselines for the WN18RR and FB15k-237 datasets are listed in Table \ref{tab:mainresult}. 
We categorize the baseline models into two main groups based on scoring functions, namely semantic matching and geometric distance scoring functions.
The models based on \textbf{S}emantic \textbf{M}atching are listed in the upper part of the table, while the \textbf{G}eometric \textbf{D}istance based methods are listed in the lower part of the table.
It is worth noting that our model's scoring function is the unique scoring function that simultaneously measures both \textbf{S}emantic and \textbf{G}eometric distances.

From Table \ref{tab:mainresult} we can clearly see that our model achieves the best results on both datasets, except for the H@1 metric on the WN18RR dataset.
Specifically, compared to the best performing of the semantic matching model, QuatRE, our model drops from 1986 to 899 on the MR metric and absolutely improves 3.4\%, 5.0\%, 3.6\% and 2.5\% on the MRR, H@10, H@3 and H@1 metrics on the WN18RR dataset.
On the FB15k-237 dataset, our model decreases from 88 to 83 on the MR metrics, and absolutely improves on the  MRR, H@10, H@3 and H@1 metrics by 1.6\%, 1.5\%, 1.4\% and 1.8\%.

Compared to the latest and best performance of the geometric distance model, TransERR, our model decreases from 1167 to 899 on the MR metric and achieves an absolute improvement of 1.8\%, 2.8\%, and 3.4\% on the MRR, H@10 and H@3 metrics on the WN18RR dataset.
On the FB15k-237 dataset, our model decreases from 125 to 83 on the MR metrics, and absolutely improves on the  MRR, H@10, H@3 and H@1 metrics by 3.6\%, 3.0\%, 3.5\% and 3.7\%, respectively.

The KGC results on WN18 and FB15k datasets are shown in Table \ref{tab:subresult}.
The Table \ref{tab:subresult} illustrates our model superiority over any previous model on the FB15k dataset. 
On the WN18 dataset, our model achieves the best results on all metrics, except for the H@1 metric which achieves second place.
In conclusion, our model not only achieves optimal results compared to semantic matching models, but also achieves competitive results compared to geometric distance models.

\begin{figure*}
\vspace{-0.5cm}
\centering 
\subfigure[1-to-N]{
\label{Fig:1-n}
\includegraphics[width=0.32\linewidth]{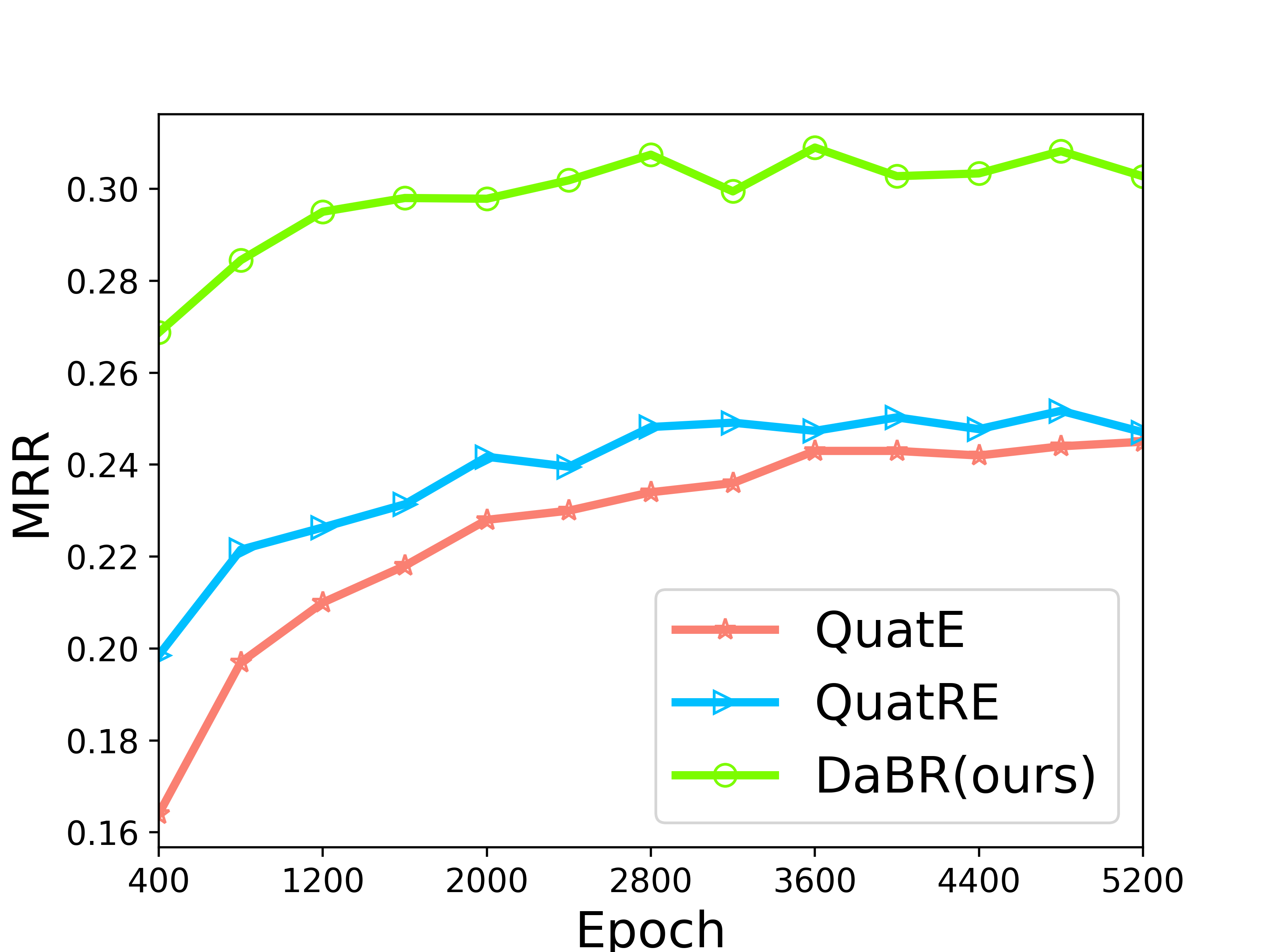}}\subfigure[N-to-1]{
\label{Fig:n-1}
\includegraphics[width=0.32\linewidth]{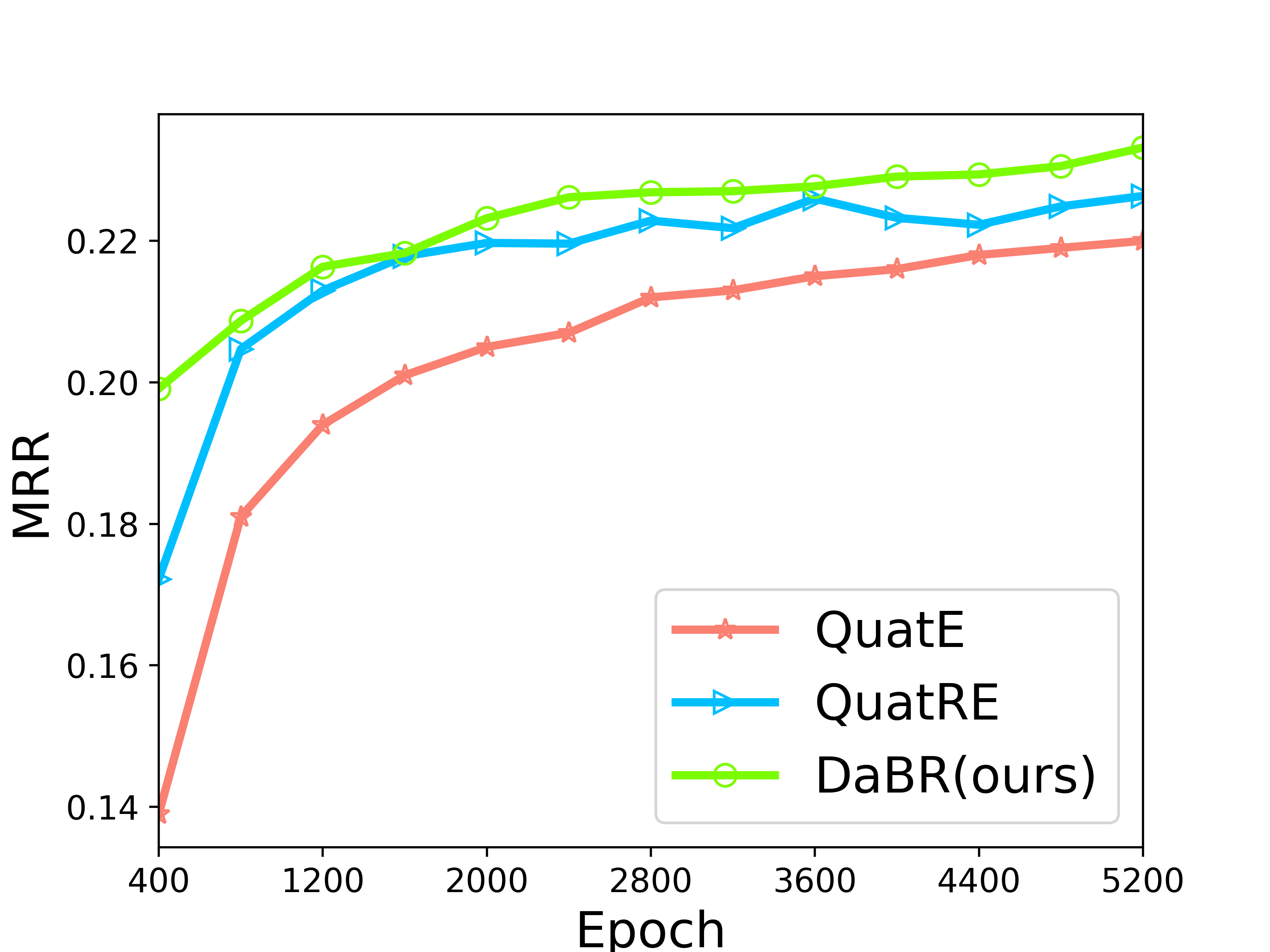}}\subfigure[N-to-N]{
\label{Fig:n-n}
\includegraphics[width=0.32\linewidth]{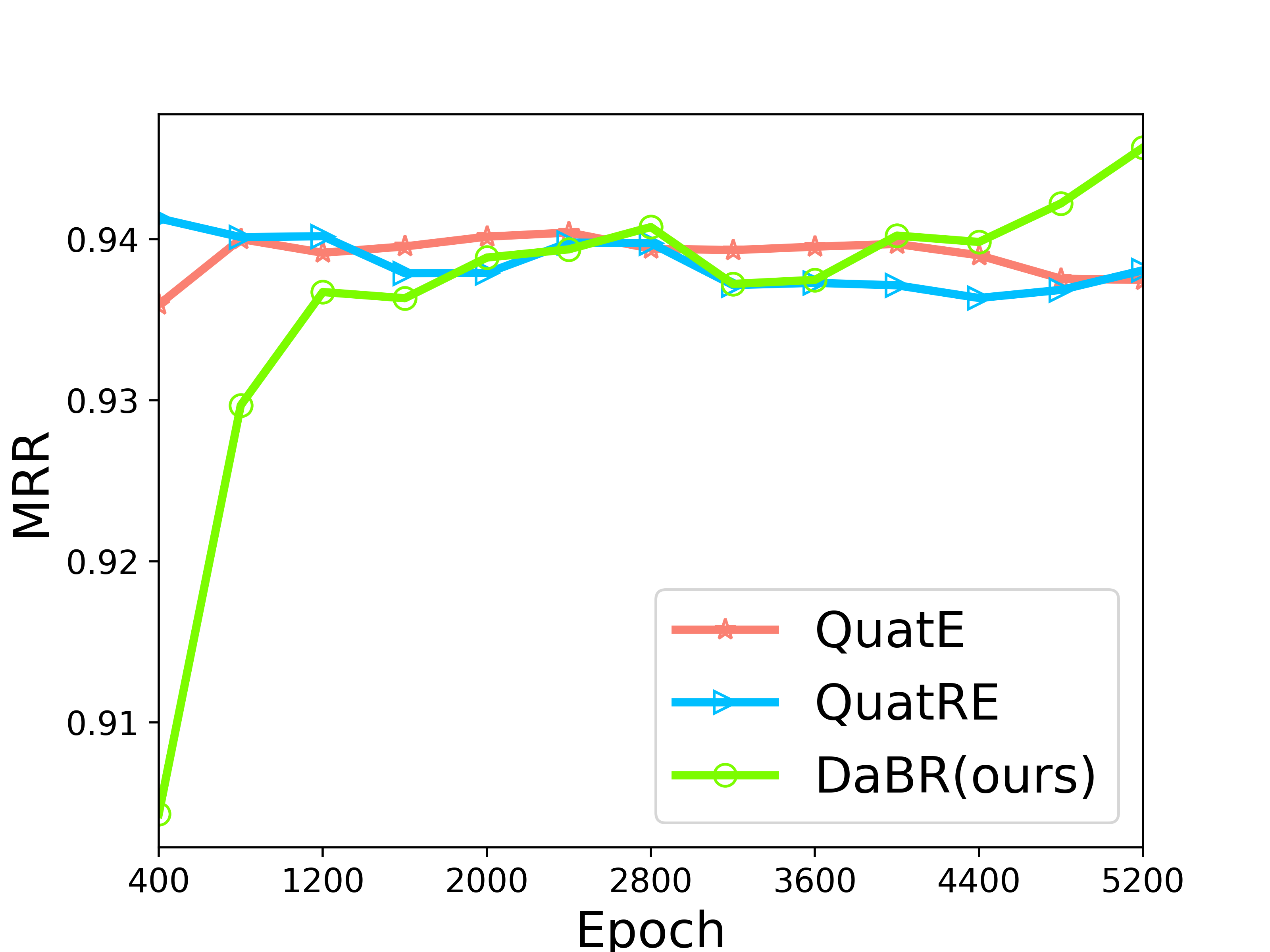}}
\caption{MRR scores for QuatE, QuatRE and our DaBR models over 0 to 5200 training epochs.}
\label{Fig:n-ncomparison}
\end{figure*}

\section{Analysis}
To demonstrate the superiority of our model, we have conducted in-depth analysis experiments from various aspects. 
The obtained experimental results and analysis are as follows:

\subsection{Ablation Analysis}
In this section, we aim to evaluate the efficacy of bidirectional rotation and distance-adaptation within our DaBR.
We have designed the following model variants: 

\textit{Variant I}: We remove the rotation of the tail entity and keep the rotation of the head entity.

\textit{Variant II}: We removed the distance-adaptation.
The DaBR degenerates into a semantic matching model.

We show the results of the ablation experiments in Table \ref{tab:ablationresult}.
From the table, we can obtain the following conclusions: 
1) The rotation of the tail entity and distance-adaptation are important parts of our model.
2) When our model removed the tail rotation, the model (i.e., \textit{Variant I}) still achieved the best results compared to the models in Table \ref{tab:mainresult} and Table \ref{tab:subresult}.
We attribute this to the fact that our model can measure both the semantics of entities and the embedding distance of entities.
3) When our model removed distance-adaptation, the model (i.e., \textit{Variant II}) performance decreased dramatically on all datasets.
It is worth noting that our model still achieves optimal results on most datasets compared to the semantic matching model on most datasets.

\subsection{Parameter Comparison Analysis}
To analyze the number of parameters compared to other models, we compared our DaBR with the best semantic matching model (QuatRE) and the best geometric distance model (TransERR).
Given the same embedding dimension $n$, QuatRE and TransERR have $(|\mathcal{E}| \times n + 3 \times |\mathcal{\mathcal{R}}| \times n)$ parameters, while our DaBR has $(|\mathcal{E}| \times n + 2 \times |\mathcal{\mathcal{R}}| \times n)$ parameters, where $\mathcal{E}$ and $\mathcal{R}$ are the entity set and relation set.
Compared to QuatRE and TransERR, our model achieves better results with fewer parameters.

\subsection{Relationship Type Analysis}
To explore the robustness of our model in the face of different relation types (one-to-many (1-to-N), many-to-one (N-to-1) and many-to-many (N-to-N)), we compared DaBR with QuatE and QuatRE in WN18R dataset.
For the results of the QuatE and QuatRE, we reproduce these models following the hyper-parameter settings of their paper.

In accordance with the calculation rules set out in \citet{bordes2013translating}, the test set of WN18RR has been divided into three categories: 1-to-N, N-to-1 and N-to-N.
The division results are shown in Appendix \ref{app:clas}, where $\eta_h$ and $\eta_t$ represent the average degree of head and tail entities, respectively.

We show the MRR scores for the QuatE, QuatRE, and DaBR models for 0 to 5200 training epochs in Figure \ref{Fig:n-ncomparison}.
This demonstrates the effectiveness of our model in modelling different types of relationships.
In particular, the model is superior in dealing with 1-to-N relationship.
``1-to-N'' means that a head entity can form a fact triplet with multiple tail entities.
We attribute this superior enhancement to the distance-adaptive embedding of our model.

\begin{figure*}[]
\centering 
\subfigure[QuatE (epoch=1)]{
\label{QuatE_1}
\includegraphics[width=0.24\linewidth]{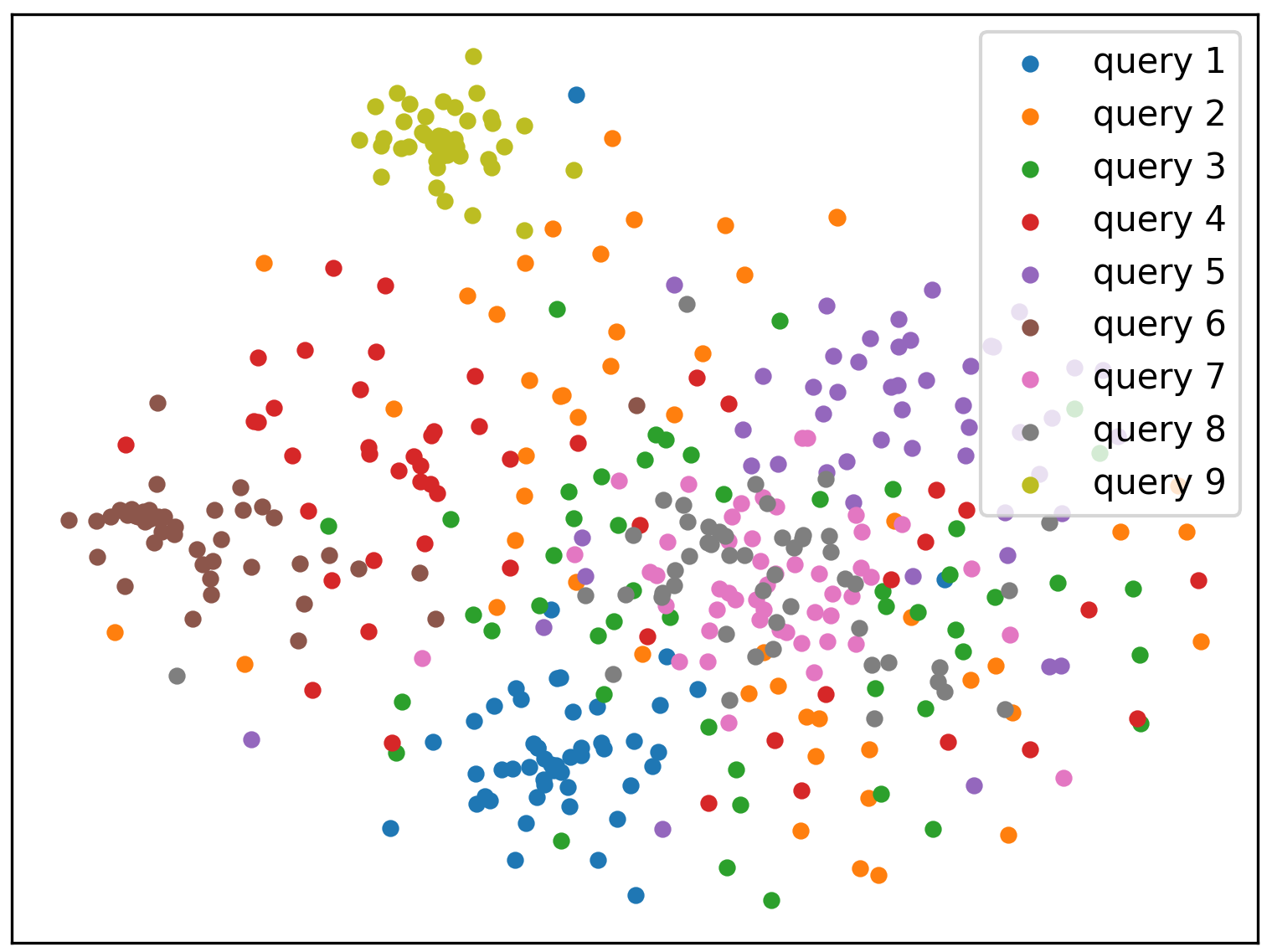}}\subfigure[QuatE (epoch=100)]{
\label{QuatE_100}
\includegraphics[width=0.24\linewidth]{fig/QuatE_100.png}}\subfigure[QuatRE (epoch=1)]{
\label{QuatRE_1}
\includegraphics[width=0.24\linewidth]{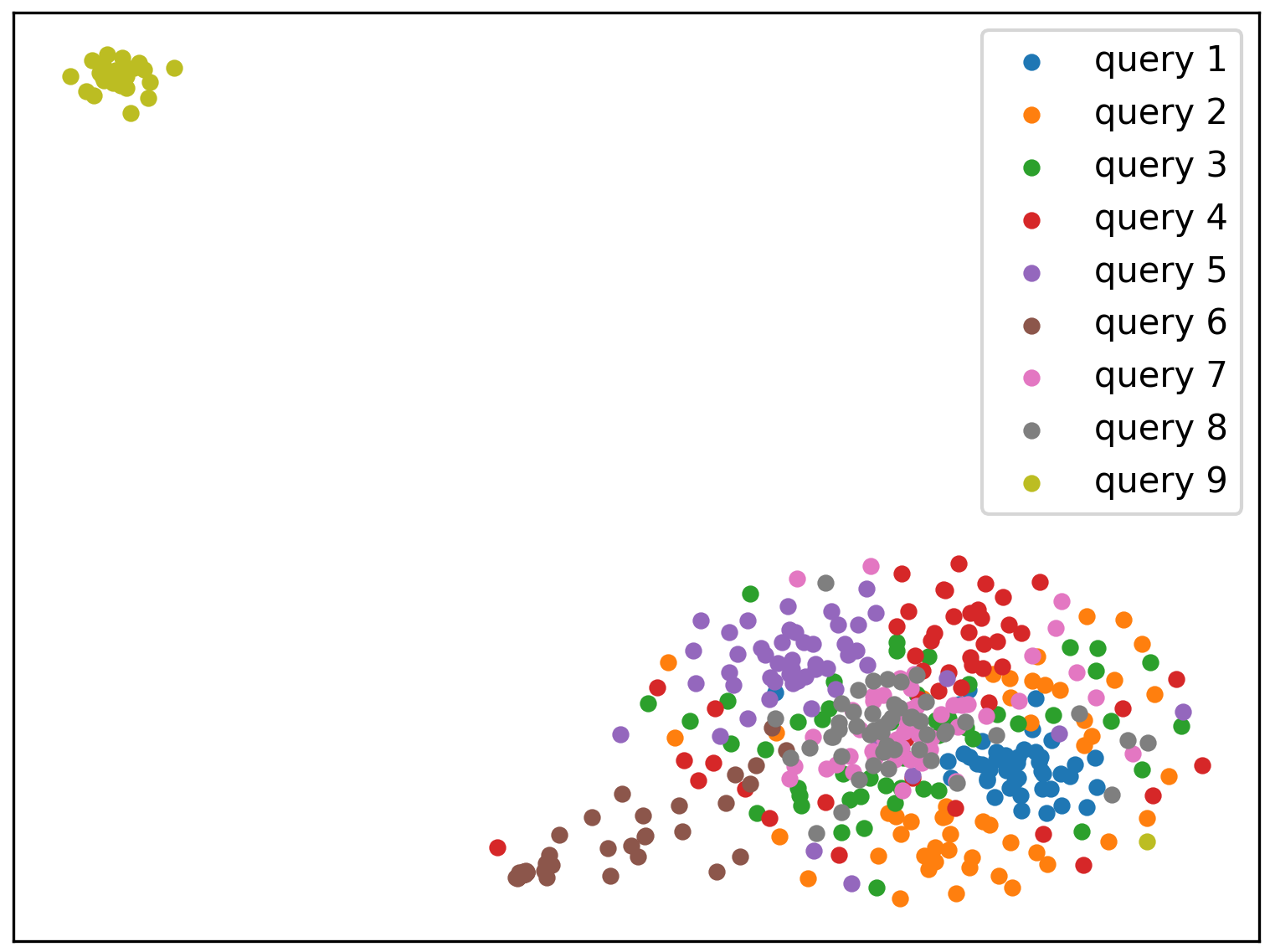}}\subfigure[QuatRE (epoch=100)]{
\label{QuatRE_100}
\includegraphics[width=0.24\linewidth]{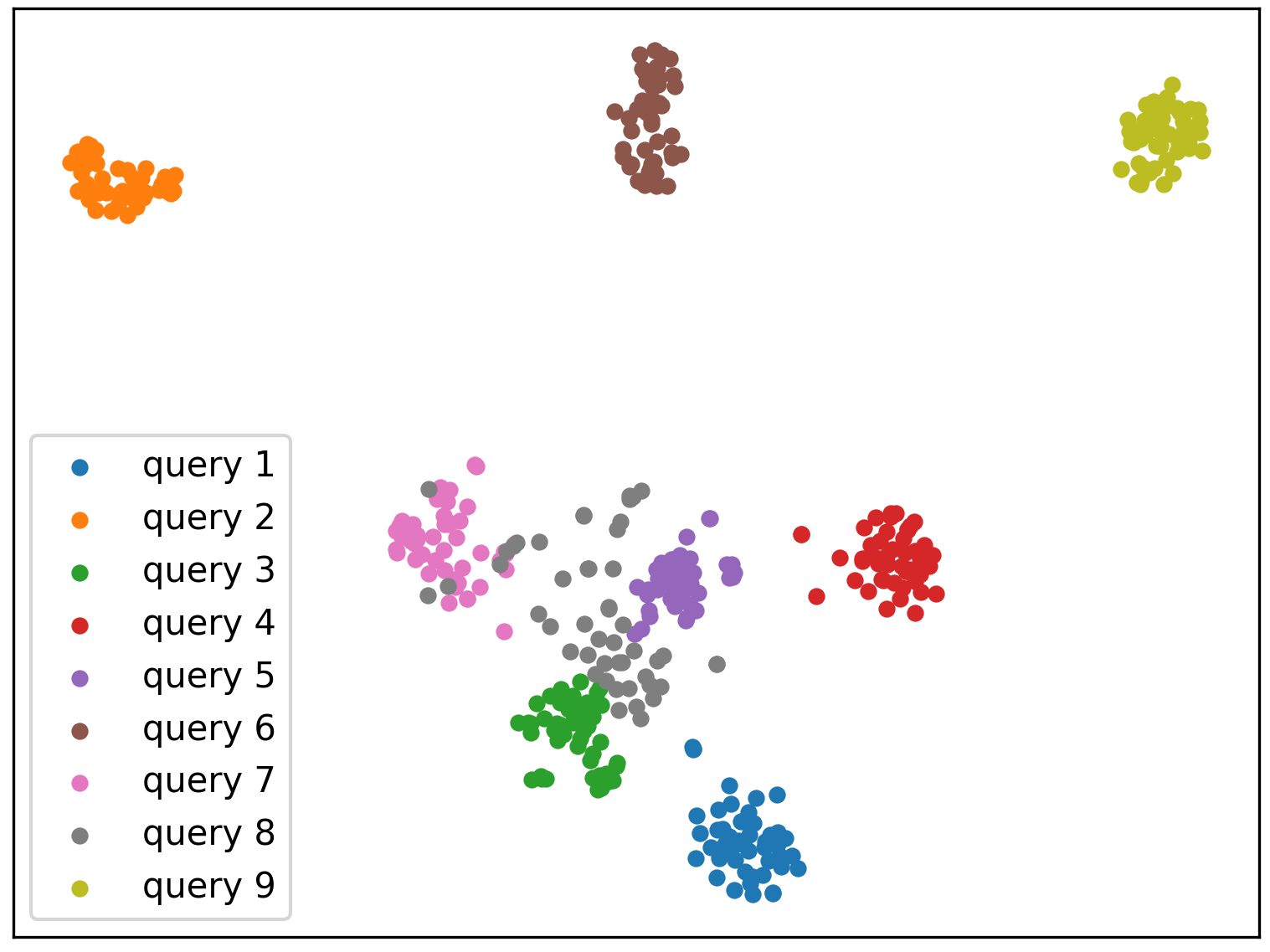}}
\subfigure[TransERR (epoch=1)]{
\label{TransERR_1}
\includegraphics[width=0.24\linewidth]{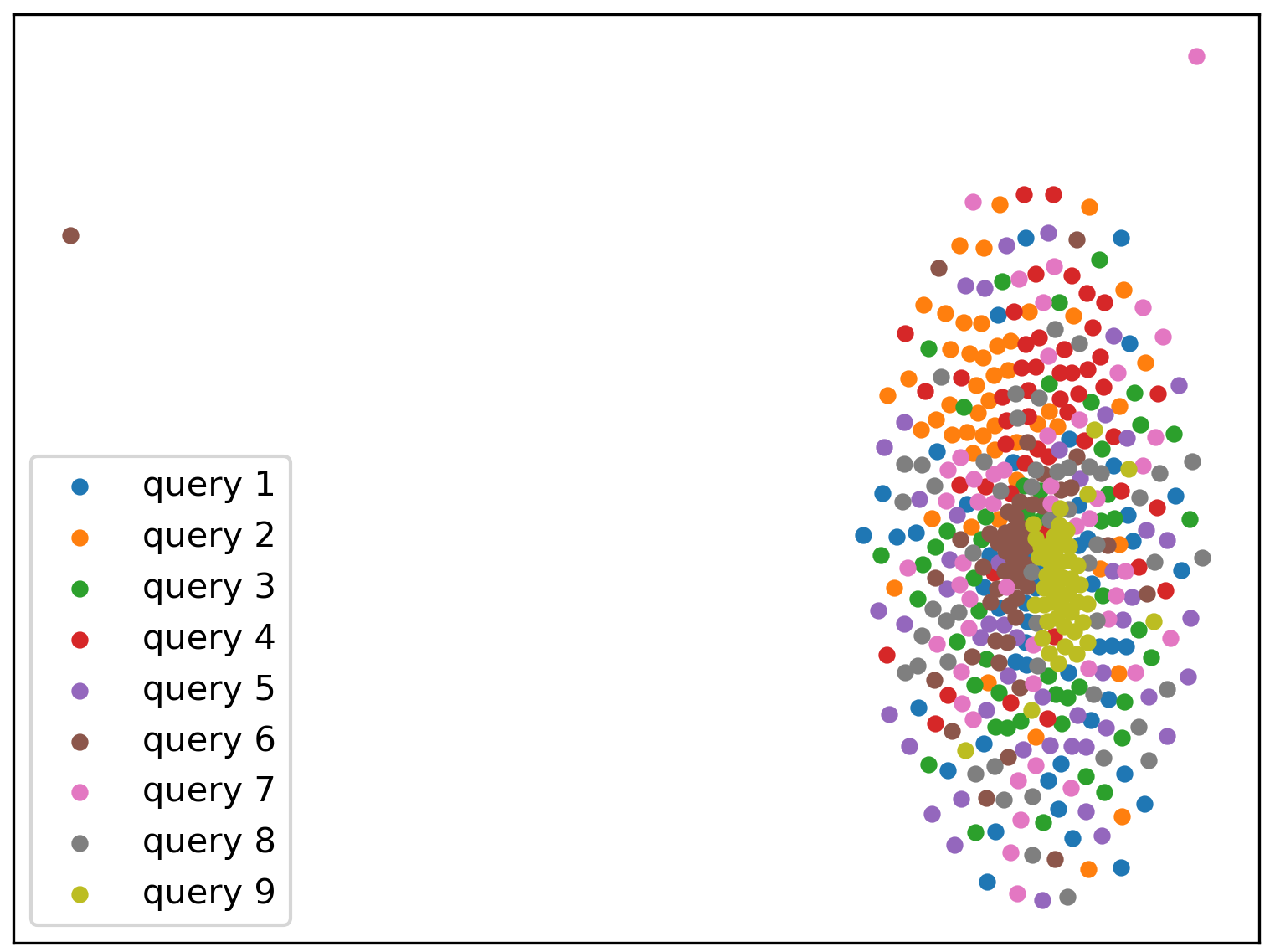}}\subfigure[TransERR (epoch=100)]{
\label{TransERR_100}
\includegraphics[width=0.24\linewidth]{fig/TransERR_100.png}}\subfigure[DaBR (epoch=1)]{
\label{DaBR_1}
\includegraphics[width=0.24\linewidth]{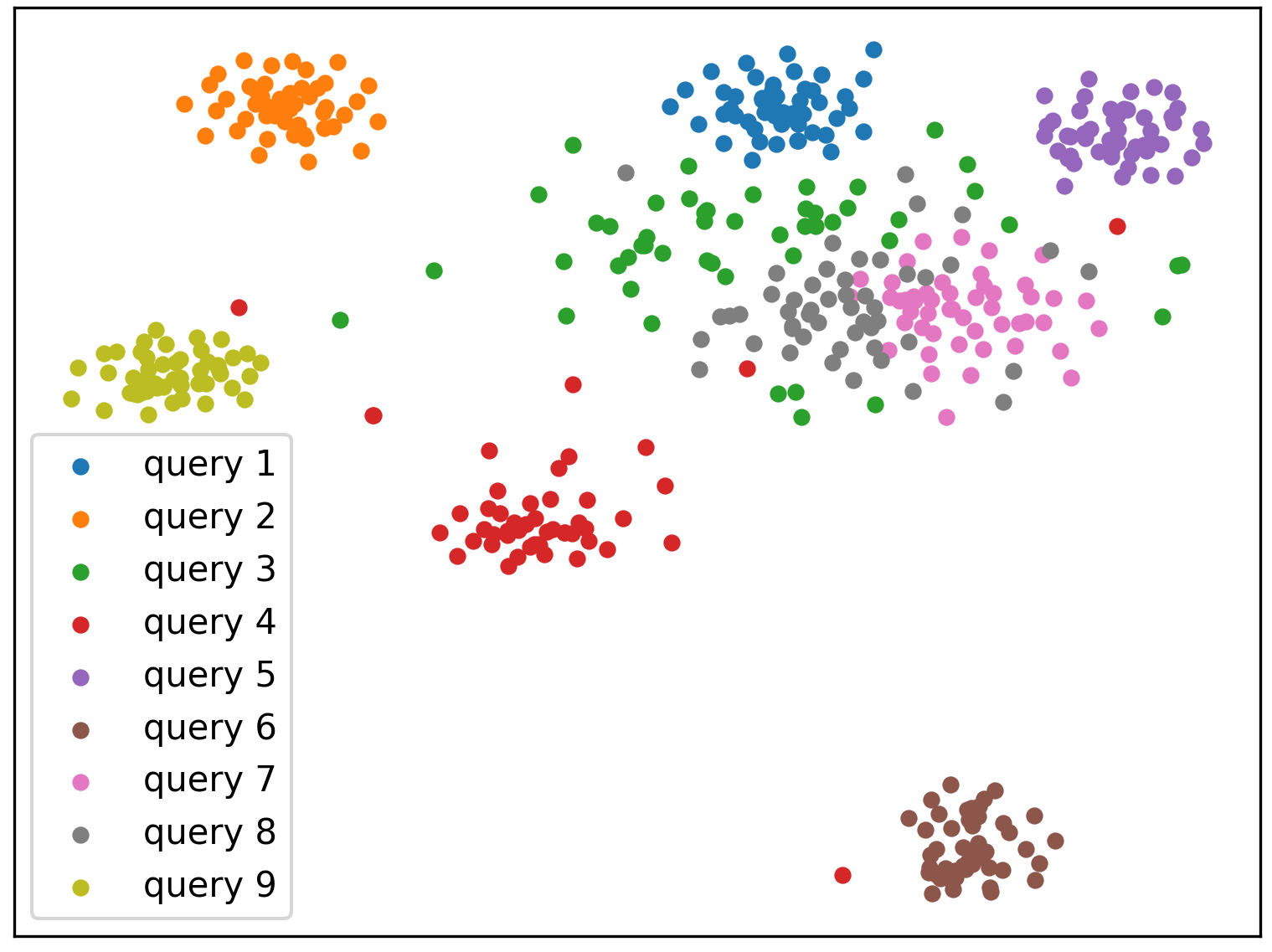}}\subfigure[DaBR (epoch=100)]{
\label{DaBR_100}
\includegraphics[width=0.24\linewidth]{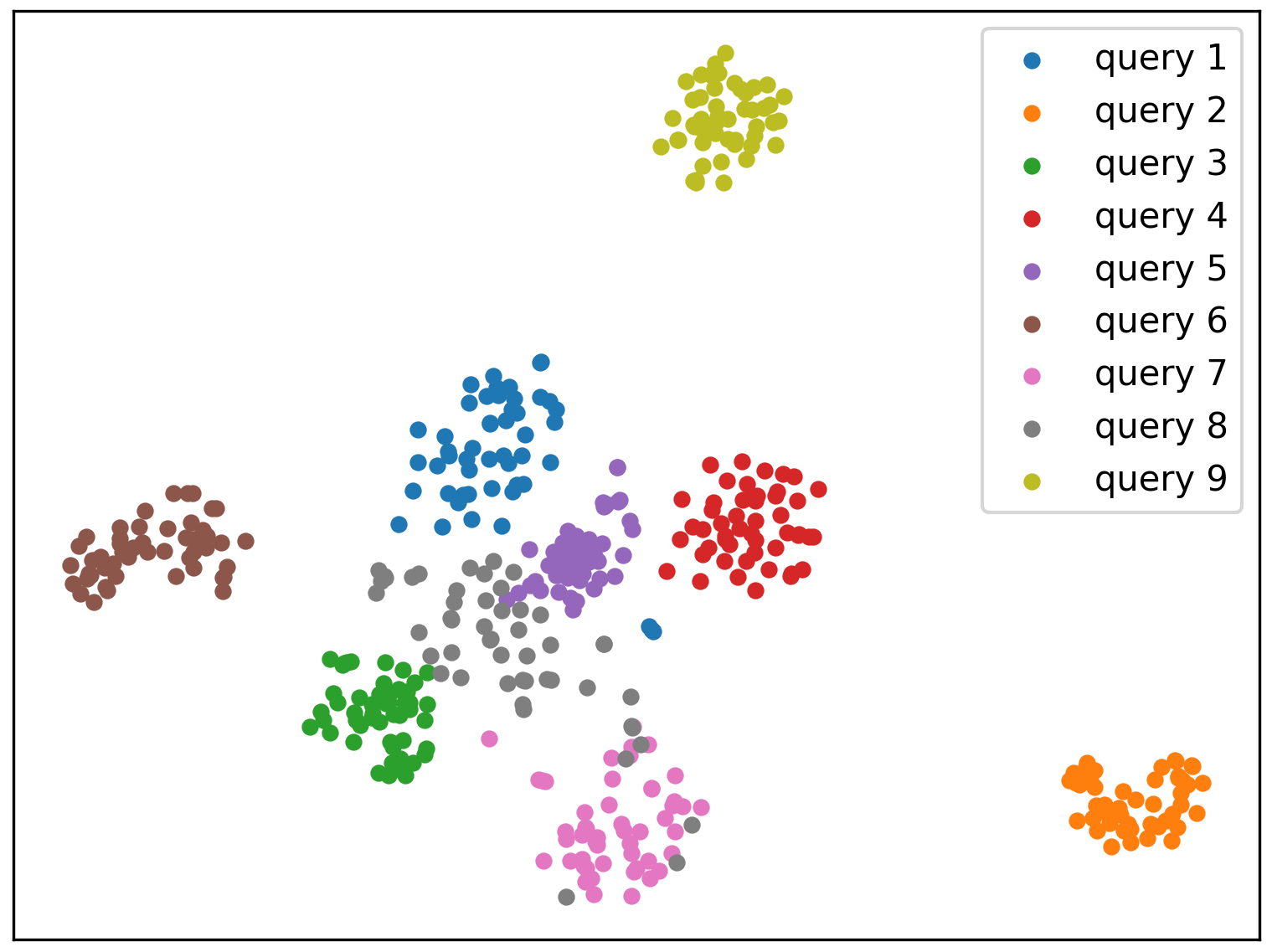}}
\caption{Visualization of the embeddings of tail entities using t-SNE. 
A point represents a tail entity. 
Points in the same color represent tail entities that have the same $(h_i, r_j )$ context.
}
\label{Fig:embedding}
\end{figure*}

\subsection{Visualization Analysis} 
\label{sec:vis}
In this section, to explore the embedding results of our model after distance adaptive embedding, we visualize the the tail entity embeddings using t-SNE \cite{JMLR:v9:vandermaaten08a}.
Suppose ($h_i$, $r_j$) is a query where $h_i$ and $r_j$ are the head entity and the relation, respectively.
If ($h_i$, $r_j$, $t_k$ ) is valid, the entity $t_k$ is the answer to query ($h_i$, $r_j$).
We selected 9 queries in FB15k-237 dataset, each of which has 50 answers. 
For more details about the 9 queries, please refer to the Appendix \ref{app:queries}.

We then use t-SNE to visualize the semantic matching models QuatE and QuatRE, the geometric distance model TransERR, and our combined semantic and geometric distance DaBR to generate the answer embeddings for epoch 1 and epoch 100, respectively.
Figure \ref{Fig:embedding} shows the visualization results\footnote{Refer to Appendix \ref{app:more} for more visualization results.}.
Each entity is represented by a 2D point and points in the same color represent tail entities with the same ($h_i$, $r_j$) context (i.e. query). 

Specifically, our model (Figure \ref{DaBR_1}) in the first epoch have demonstrated better embedding compared to QuatE, QuatRE and TransERR. 
At epoch 100, our model (Figure \ref{DaBR_100}) show clear inter-cluster separability, with entities within each cluster (intra-cluster) being well-separated from one another.

However, the semantic matching model QuatE (Figure \ref{QuatE_100}) and QuatRE (Figure \ref{QuatRE_100}) heavily overlap entities within clusters despite inter-cluster separability.
The geometric distance model TransERR (Figure \ref{TransERR_100}) clusters are indistinguishable from each other and entities within the clusters (intra-clusters) are distinguishable.

\begin{table}[h]
\centering
\begin{tabular}{ccc}
\hline
 \multicolumn{1}{c}{\textbf{Model}}   & intra-cluster  & inter-cluster        \\ 
\hline
QuatE         &     & $\checkmark $          \\ 
QuatRE         &     & $\checkmark $          \\ 
TransERR         & $\checkmark $      &            \\ 
DaBR        & $\checkmark $     & $\checkmark $          \\ 
\hline
\end{tabular}
\caption{$\checkmark $ indicates a separable ability.
}
\label{tab:sep}
\end{table}

Table \ref{tab:sep} summarizes our analysis above, which we attribute to the fact that our model combines semantic matching with entity geometric distance to better measure the plausibility of triplets.


\begin{figure}
\centering 
\subfigure[DaBR (with)]{
\label{Fig:with}
\includegraphics[width=0.48\linewidth]{fig/ours_1.png}}\subfigure[DaBR (without)]{
\label{Fig:without}
\includegraphics[width=0.48\linewidth]{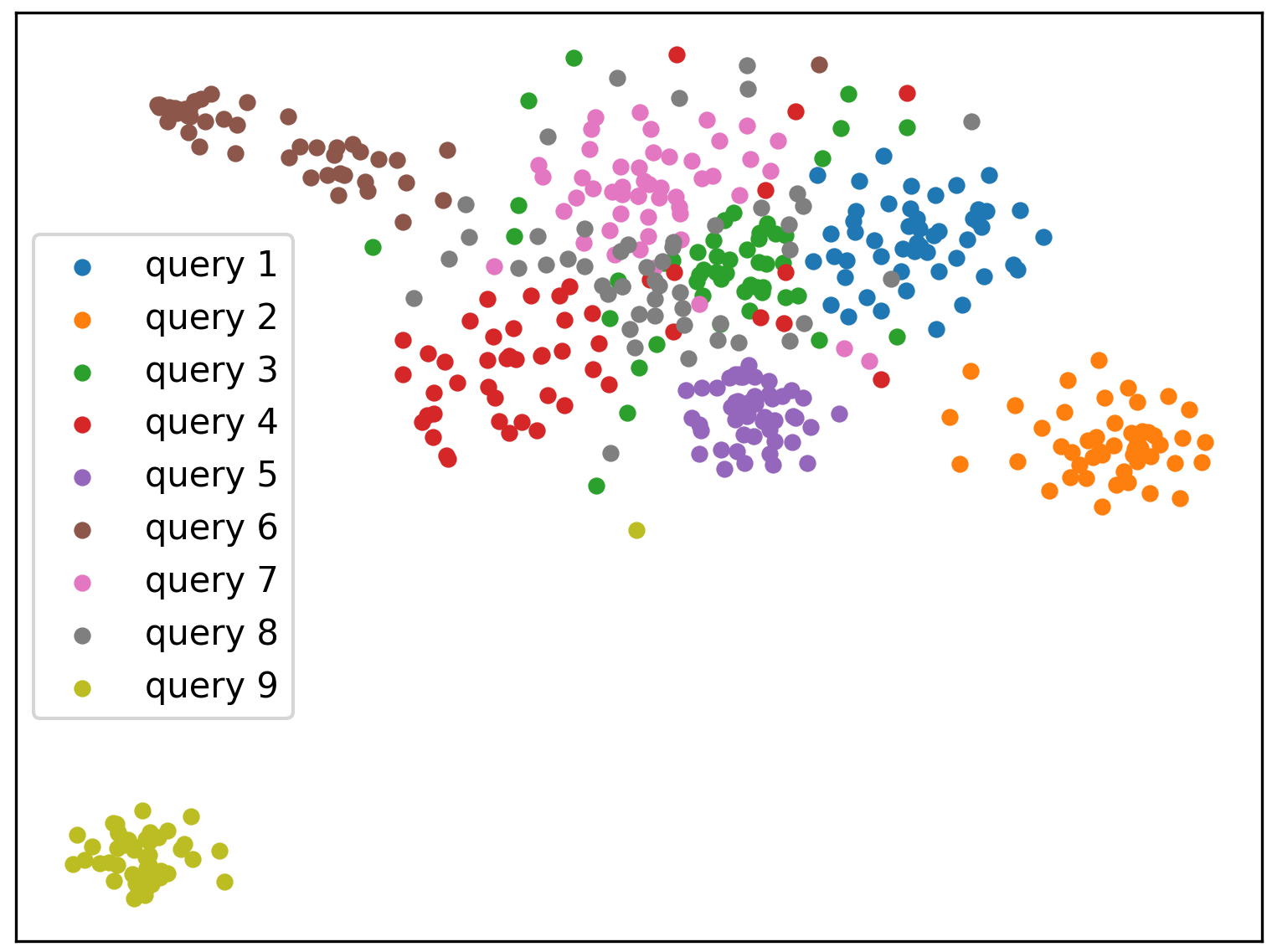}}
\caption{DaBR with distance-adaptation and without.}
\label{Fig:ablation}
\end{figure}

\subsection{Visualization Ablation Analysis}
In Figure \ref{Fig:ablation}, we visualize that our model removes the distance adaptive embedding in the first epoch.
We can find that the visualization without the distance adaptive embedding (Figure \ref{Fig:without}) is worse than the with one (Figure \ref{Fig:with}).
By visualizing the ablation experiments, we can further illustrate the advantage of distance adaptive embedding.

\section{Conclusion}

We note that existing quaternion models based on semantic matching diminishes the separability of entities, while the distance scoring function weakens the semantics of entities.
To address this issue, we propose a novel quaternion knowledge graph embedding model.
By combining semantic matching with entity geometric distance, our model provides a robust and comprehensive framework for knowledge graph embedding.
We provide mathematical proofs to demonstrate our model can handle complex logical relationships.
Visualization results show that our model can learn the geometric distance property between entities to achieve both inter-cluster and intra-cluster separability.

\section*{Limitations}
The H@1 metric performance of our model on the WN18 and WN18RR datasets is not optimal.
In addition, like most knowledge graph embedding models, our model is unable to predict new entities that do not exist in the training data.










\section*{Acknowledgements}
This work is supported by National Natural Science Foundation of China (No.62066033);
Inner Mongolia Natural Science Foundation (Nos.2024MS06013, 2022JQ05); 
Inner Mongolia Autonomous Region Science and Technology Programme Project (Nos.2023YFSW0001, 2022YFDZ0059, 2021GG0158);
We also thank all anonymous reviewers for their insightful comments.


\bibliography{custom}

\clearpage
\appendix
\section*{Appendix}

\begin{table*}[t]
\renewcommand\arraystretch{1.1}
\centering
\resizebox{\textwidth}{!}{
\begin{tabular}{cl}
\hline
\textbf{Index} & \textbf{Query}           \\ 
\hline
1 &(political drama, $/$media\_common$/$netflix\_genre$/$titles, ?)    \\
2 &(Academy Award for Best Original Song, $/$award$/$award\_category$/$winners.$/$award$/$award\_honor$/$ceremony, ?) \\
3 &(Germany, $/$location$/$location$/$contains, ?) \\
4 &(Master's Degree, $/$education$/$educational\_degree$/$people\_with\_this\_degree.$/$education$/$education$/$major\_field\_of\_study, ?) \\
5 &(broccoli, $/$food$/$food$/$nutrients.$/$food$/$nutrition\_fact$/$nutrient, ?) \\
6 &(shooting sport, $/$olympics$/$olympic\_sport$/$athletes.$/$olympics$/$olympic\_athlete\_affiliation$/$country,?) \\
7 &(synthpop, $/$music$/$genre$/$artists, ?) \\
8 &(Italian American, $/$people$/$ethnicity$/$people, ?) \\
9 &(organ, $/$music$/$performance\_role$/$track\_performances.$/$music$/$track\_contribution$/$role, ?) \\
\hline
\end{tabular}
}
\caption{The queries in t-SNE visualizations.}
\label{tab:queries}
\end{table*}

\section{Proof}
\label{app:proof}
Given $\mathbf{h} = a_h + b_h \mathbf{i} + c_h \mathbf{j} + d_h \mathbf{k}, \mathbf{r} = p + q \mathbf{i} + u \mathbf{j} + v \mathbf{k}, \mathbf{t} = a_t + b_t \mathbf{i} + c_t \mathbf{j} + d_t \mathbf{k}$, where $\mathbf{r}$ is a unit quaternion after normalization operation.
We can make $\lambda=0$ and then our scoring function can be simplified as follows:
\begin{equation}
\begin{aligned}
    &\phi(h,r,t)=\mathbf{h}\otimes\mathbf{r} \cdot \mathbf{t}\otimes \mathbf{\bar{r}} \\ 
    & = [ (a_h \circ p - b_h \circ q - c_h \circ u - d_h \circ v) \\
    & + (a_h \circ q + b_h \circ p + c_h \circ v - d_h \circ u) \mathbf{i}  \\
    & + (a_h \circ u - b_h \circ v + c_h \circ p + d_h \circ q) \mathbf{j} \\
    & + (a_h \circ v + b_h \circ u - c_h \circ q + d_h \circ p) \mathbf{k} ] \\
    &\cdot [ (a_t \circ p + b_t \circ q + c_t \circ u + d_t \circ v) \\
    & + (- a_t \circ q + b_t \circ p - c_t \circ v + d_t \circ u) \mathbf{i}  \\
    & + (- a_t \circ u + b_t \circ v + c_t \circ p - d_t \circ q) \mathbf{j} \\
    & + (- a_t \circ v - b_t \circ u + c_t \circ q + d_t \circ p) \mathbf{k} ] \\
\end{aligned}
\end{equation}
where $\otimes$ is the Hamilton product, $\circ$ denotes the element-wise product, and ``$\cdot$'' is the inner product.

\subsection{Proof of Symmetry pattern}
\label{pr1}
In order to prove the symmetry pattern, we need to prove the following equality:
\begin{equation}
    \mathbf{h}\otimes\mathbf{r} \cdot \mathbf{t}\otimes \mathbf{\bar{r}} = \mathbf{t}\otimes\mathbf{r} \cdot \mathbf{h}\otimes \mathbf{\bar{r}}.
\end{equation}
The symmetry property of DaBR can be proved by setting the imaginary parts of $\mathbf{r}$ to zero.

\subsection{Proof of Antisymmetry pattern}
\label{pr2}
In order to prove the antisymmetry pattern, we need to prove the following inequality when imaginary components are nonzero:
\begin{equation}
    \mathbf{h}\otimes\mathbf{r} \cdot \mathbf{t}\otimes \mathbf{\bar{r}} \neq \mathbf{t}\otimes\mathbf{r} \cdot \mathbf{h}\otimes \mathbf{\bar{r}}.
\end{equation}
We expand the right term:
\begin{equation}
\begin{aligned}
    &\mathbf{t}\otimes\mathbf{r} \cdot \mathbf{h}\otimes \mathbf{\bar{r}} \\ 
    & = [ (a_t \circ p - b_t \circ q - c_t \circ u - d_t \circ v) \\
    & + (a_t \circ q + b_t \circ p + c_t \circ v - d_t \circ u) \mathbf{i}  \\
    & + (a_t \circ u - b_t \circ v + c_t \circ p + d_t \circ q) \mathbf{j} \\
    & + (a_t \circ v + b_t \circ u - c_t \circ q + d_t \circ p) \mathbf{k} ] \\
    &\cdot [ (a_h \circ p + b_h \circ q + c_h \circ u + d_h \circ v) \\
    & + (- a_h \circ q + b_h \circ p - c_h \circ v + d_h \circ u) \mathbf{i}  \\
    & + (- a_h \circ u + b_h \circ v + c_h \circ p - d_h \circ q) \mathbf{j} \\
    & + (- a_h \circ v - b_h \circ u + c_h \circ q + d_h \circ p) \mathbf{k} ]. \\
\end{aligned}
\end{equation}
We can easily see that those two terms are not equal as the signs for some terms are not the same.

\subsection{Proof of Inversion pattern}
\label{pr3}
To prove the inversion pattern, we need to prove that:
\begin{equation}
    \mathbf{h}\otimes\mathbf{r} \cdot \mathbf{t}\otimes \mathbf{\bar{r}} = \mathbf{t}\otimes\mathbf{\bar{r}} \cdot \mathbf{h}\otimes \mathbf{\bar{r}}^{-1}.
\end{equation}
We expand the right term:
\begin{equation}
\begin{aligned}
    &\mathbf{t}\otimes\mathbf{\bar{r}} \cdot \mathbf{h}\otimes \mathbf{\bar{r}}^{-1} \\ 
    & = \mathbf{t}\otimes\mathbf{\bar{r}} \cdot \mathbf{h}\otimes \mathbf{r}\\
    & = [ (a_t \circ p + b_t \circ q + c_t \circ u + d_t \circ v) \\
    & + (- a_t \circ q + b_t \circ p - c_t \circ v + d_t \circ u) \mathbf{i}  \\
    & + (- a_t \circ u + b_t \circ v + c_t \circ p - d_t \circ q) \mathbf{j} \\
    & + (- a_t \circ v - b_t \circ u + c_t \circ q + d_t \circ p) \mathbf{k} ] \\
    &\cdot [ (a_h \circ p - b_h \circ q - c_h \circ u - d_h \circ v) \\
    & + (a_h \circ q + b_h \circ p + c_h \circ v - d_h \circ u) \mathbf{i}  \\
    & + (a_h \circ u - b_h \circ v + c_h \circ p + d_h \circ q) \mathbf{j} \\
    & + (a_h \circ v + b_h \circ u - c_h \circ q + d_h \circ p) \mathbf{k} ]. \\
\end{aligned}
\end{equation}
We can easily check the equality of these two terms.
Since $\mathbf{r}$ is a unit quaternion, we have $\mathbf{r}^{-1} = \mathbf{\bar{r}}$.

\subsection{Proof of Composition pattern}
\label{pr4}
For composition relationships, we can get that:
\begin{equation}
\begin{aligned}
    &(\mathbf{h}\otimes\mathbf{r_2})\otimes\mathbf{r_3} \cdot (\mathbf{t}\otimes \mathbf{\bar{r}_2})\otimes\mathbf{\bar{r}_3} \\
    &=\mathbf{h}\otimes(\mathbf{r_2}\otimes\mathbf{r_3}) \cdot \mathbf{t}\otimes (\mathbf{\bar{r}_2}\otimes\mathbf{\bar{r}_3}) \\
    &=\mathbf{h}\otimes\mathbf{r_1} \cdot \mathbf{t}\otimes \mathbf{\bar{r}_1}
\end{aligned}
\end{equation}


\begin{table}[h]
\renewcommand\arraystretch{1}
\centering
\setlength{\tabcolsep}{3.3pt}
\begin{tabular}{cccccc}
\hline
\textbf{Dataset} & \#Ent &  \#Rel&  \#Train  & \#Valid   & \#Test                     \\ 
\hline
WN18RR       & $40k$        & $11$       & $86k$       & $3k$     &$3k$    \\
FB15k-237    & $14k$        & $237$      & $272k$      & $17k$    &$20k$  \\
WN18         & $40k$        & $18$       & $141k$      & $5k$      &$5k$     \\
FB15k        & $14k$        & $1345$    & $483k$      & $50k$    &$59k$     \\
\hline
\end{tabular}
\caption{Dataset statistics on four datasets.}
\label{tab:datastat}
\end{table}

\section{Dataset statistics}
\label{app:datasta}
The detailed statistics of the four standard datasets are shown in Table \ref{tab:datastat}.

\begin{table}
\centering
\setlength{\tabcolsep}{4.5pt}
\begin{tabular}{cccccc}
\hline
\textbf{Dataset} & $lr$ &  $neg$  &  $dim$  & $\eta_1$  & $\eta_2$                     \\ 
\hline
WN18RR   &0.1 &5  &500  &0.5 &0.01  \\
FB15k-237 &0.05  &10  &500  &0.5 &0.01   \\
WN18    &0.05  &5  &300  &0.05 &0.01      \\
FB15k   &0.02  &10  &400  &0.05 &0.01     \\
\hline
\end{tabular}
\caption{Optimal hyper-parameters for our DaBR on each dataset.}
\label{tab:para}
\end{table}

\begin{table}
\renewcommand\arraystretch{1}
\centering
\begin{tabular}{cccc}
\hline
 \multicolumn{1}{c}{\textbf{Category}}        & $\eta_h$   & $\eta_t$     & \#triplets        \\ 
\hline
\multicolumn{1}{c}{1-to-N}         & $<1.5$     & $>1.5$       & 475    \\ 
\multicolumn{1}{c}{N-to-1}         & $>1.5$     & $<1.5$       & 1487    \\ 
\multicolumn{1}{c}{N-to-N}        & $>1.5$     & $>1.5$       & 1130    \\ 
\hline
\end{tabular}
\caption{Classification rules and classification results for WN18RR.
The last column is the number after division.
}
\label{tab:categories}
\end{table}

\section{Optimal hyper-parameters}
\label{app:optimal}
Table \ref{tab:para} shows the optimal hyperparameter settings for our model on the four benchmark datasets. The optimal parameters come from the highest scores of our model on the validation dataset.

\section{Classification rules}
\label{app:clas}
The classification rules and classification results for WN18RR dataset in the Table \ref{tab:categories}.

\section{The queries in t-SNE visualization}
\label{app:queries}
In Table \ref{tab:queries}, we list the nine queries used in the t-SNE visualization (Section \ref{sec:vis} in the main text). 
Note that a query is represented as $(h, r, ?)$, where $h$ denotes the head entity and $r$ denotes the relation.

\section{More visualization results}
\label{app:more}
\begin{figure*}[t]
\centering 
\subfigure[QuatE (epoch=1)]{
\includegraphics[width=0.32\linewidth]{fig/QuatE_1.png}}\subfigure[QuatE (epoch=50)]{
\includegraphics[width=0.32\linewidth]{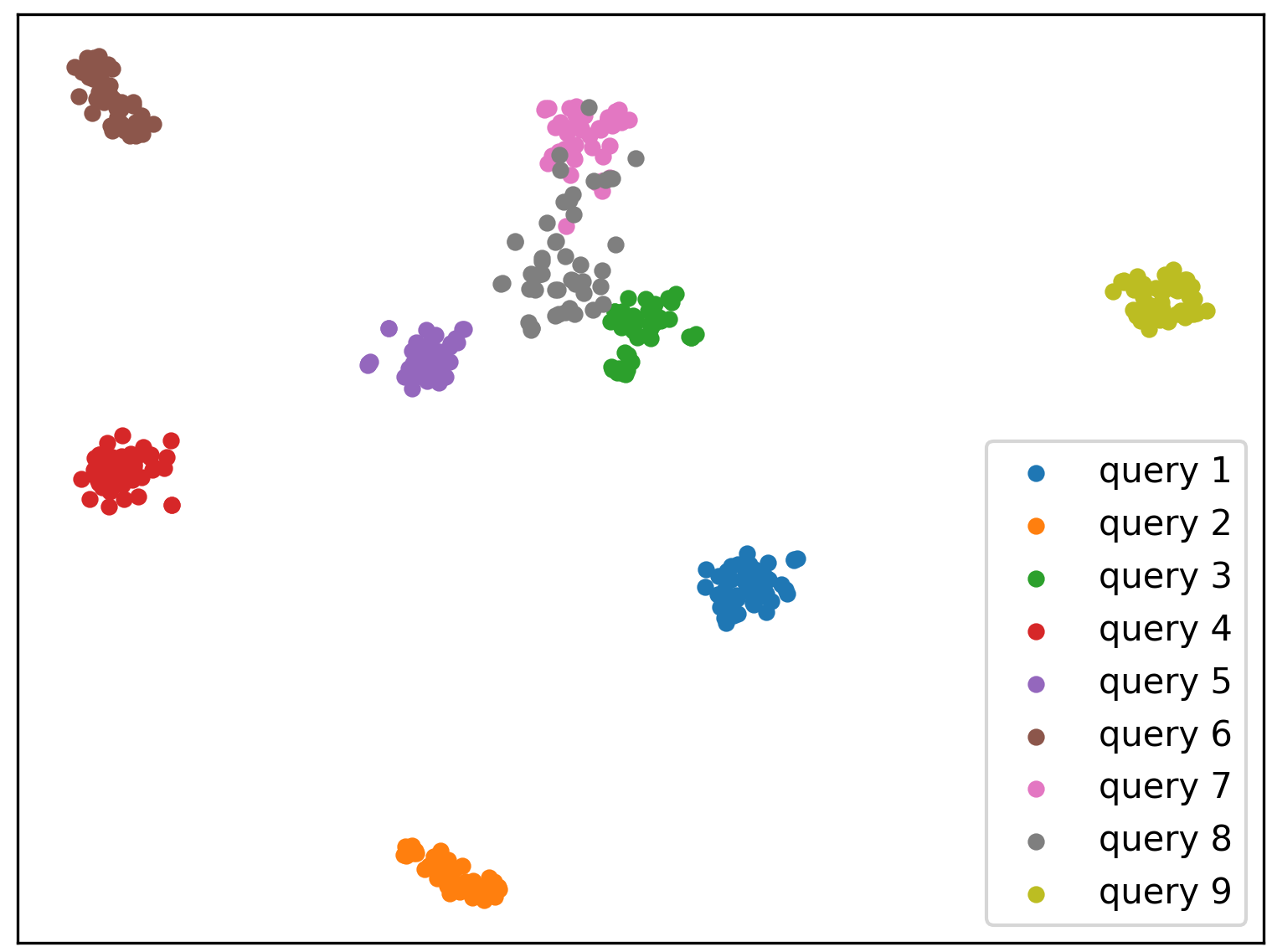}}\subfigure[QuatE (epoch=100)]{
\includegraphics[width=0.32\linewidth]{fig/QuatE_100.png}}
\subfigure[QuatRE (epoch=1)]{
\includegraphics[width=0.32\linewidth]{fig/QuatRE_1.png}}\subfigure[QuatRE (epoch=50)]{
\includegraphics[width=0.32\linewidth]{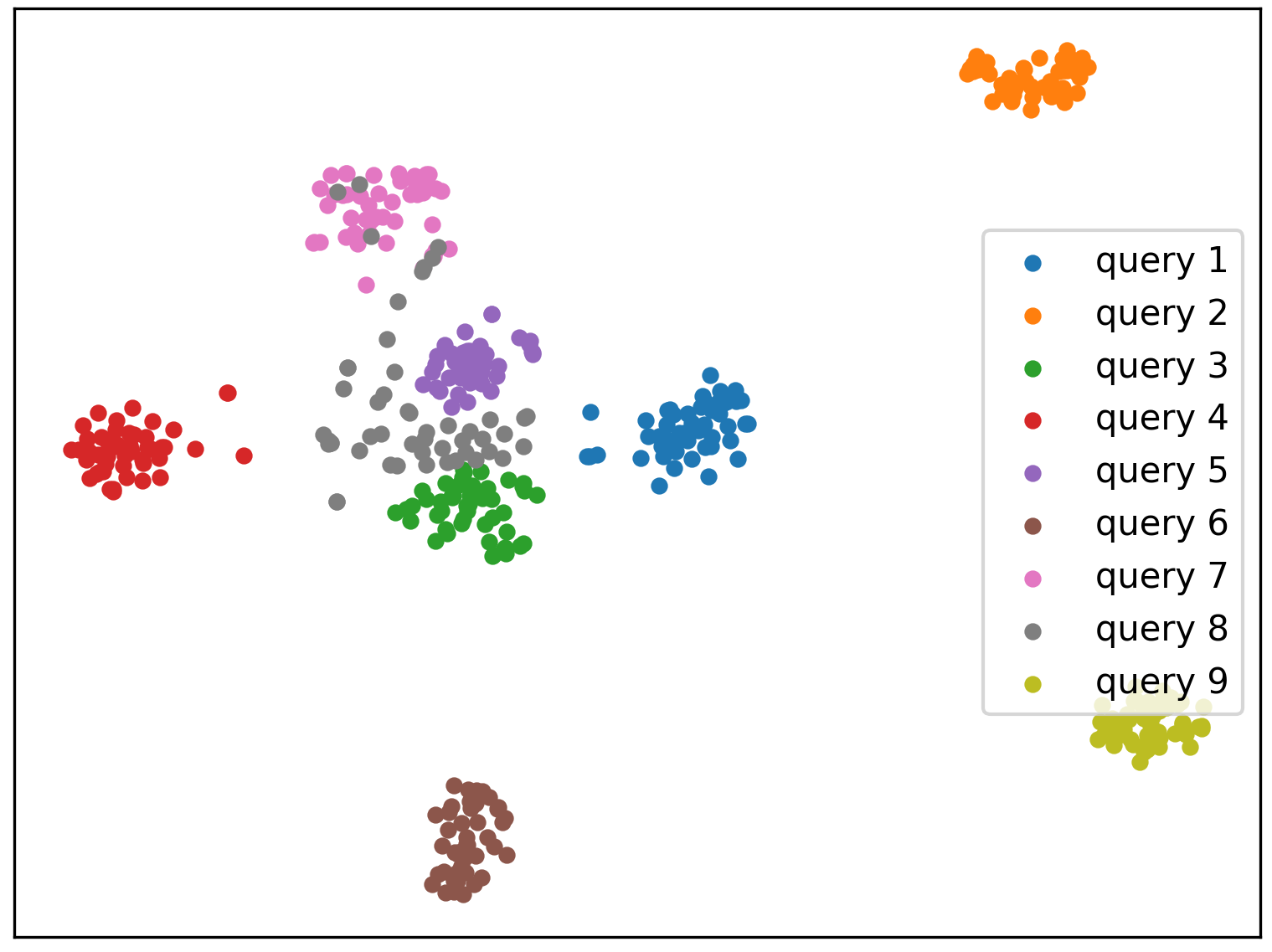}}\subfigure[QuatRE (epoch=100)]{
\includegraphics[width=0.32\linewidth]{fig/QuatRE_100.png}}
\subfigure[TransERR (epoch=1)]{
\includegraphics[width=0.32\linewidth]{fig/TransERR_1.png}}\subfigure[TransERR (epoch=50)]{
\includegraphics[width=0.32\linewidth]{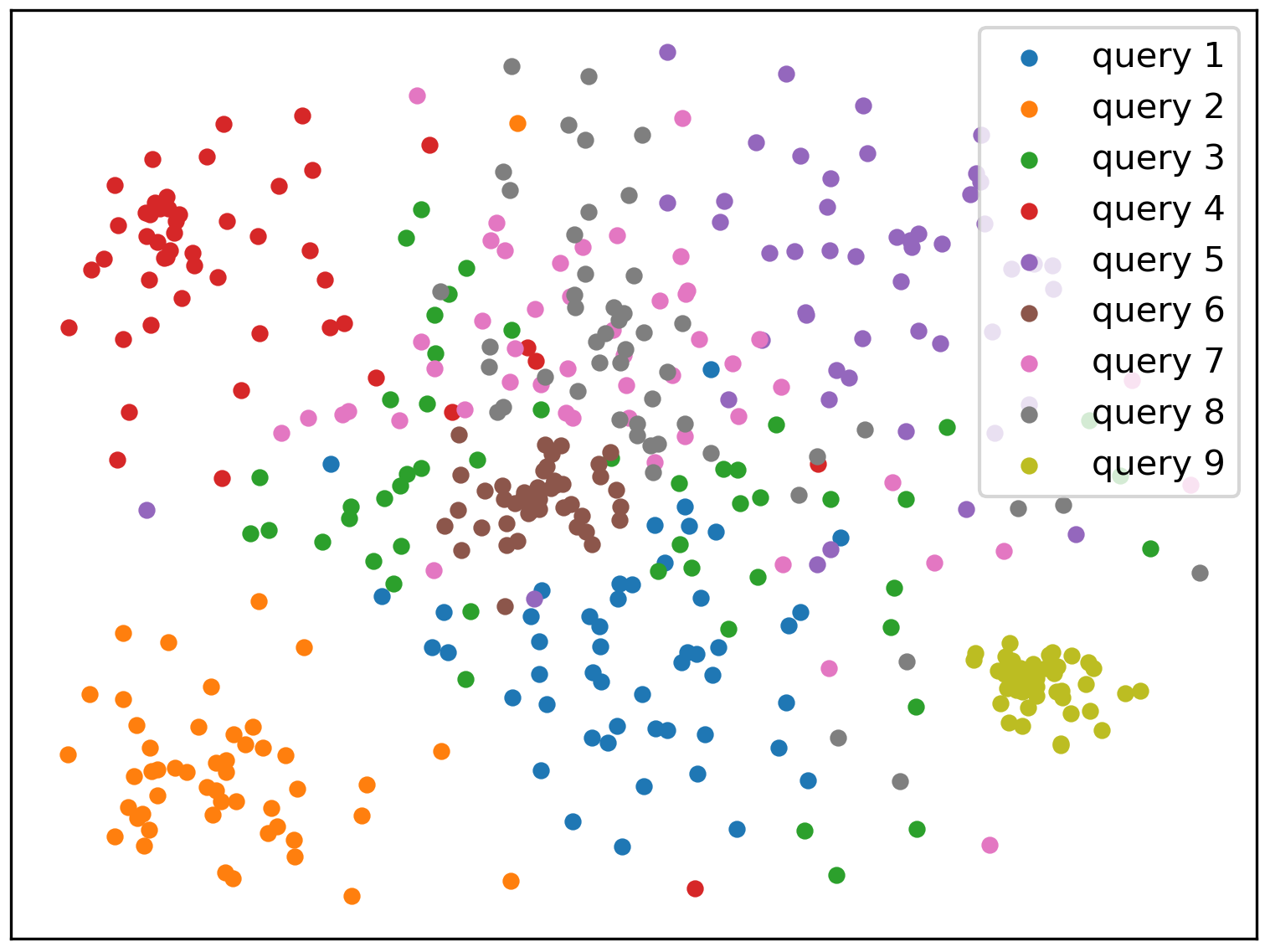}}\subfigure[TransERR (epoch=100)]{
\includegraphics[width=0.32\linewidth]{fig/TransERR_100.png}}
\subfigure[DaBR (epoch=1)]{
\includegraphics[width=0.32\linewidth]{fig/ours_1.png}}\subfigure[DaBR (epoch=50)]{
\includegraphics[width=0.32\linewidth]{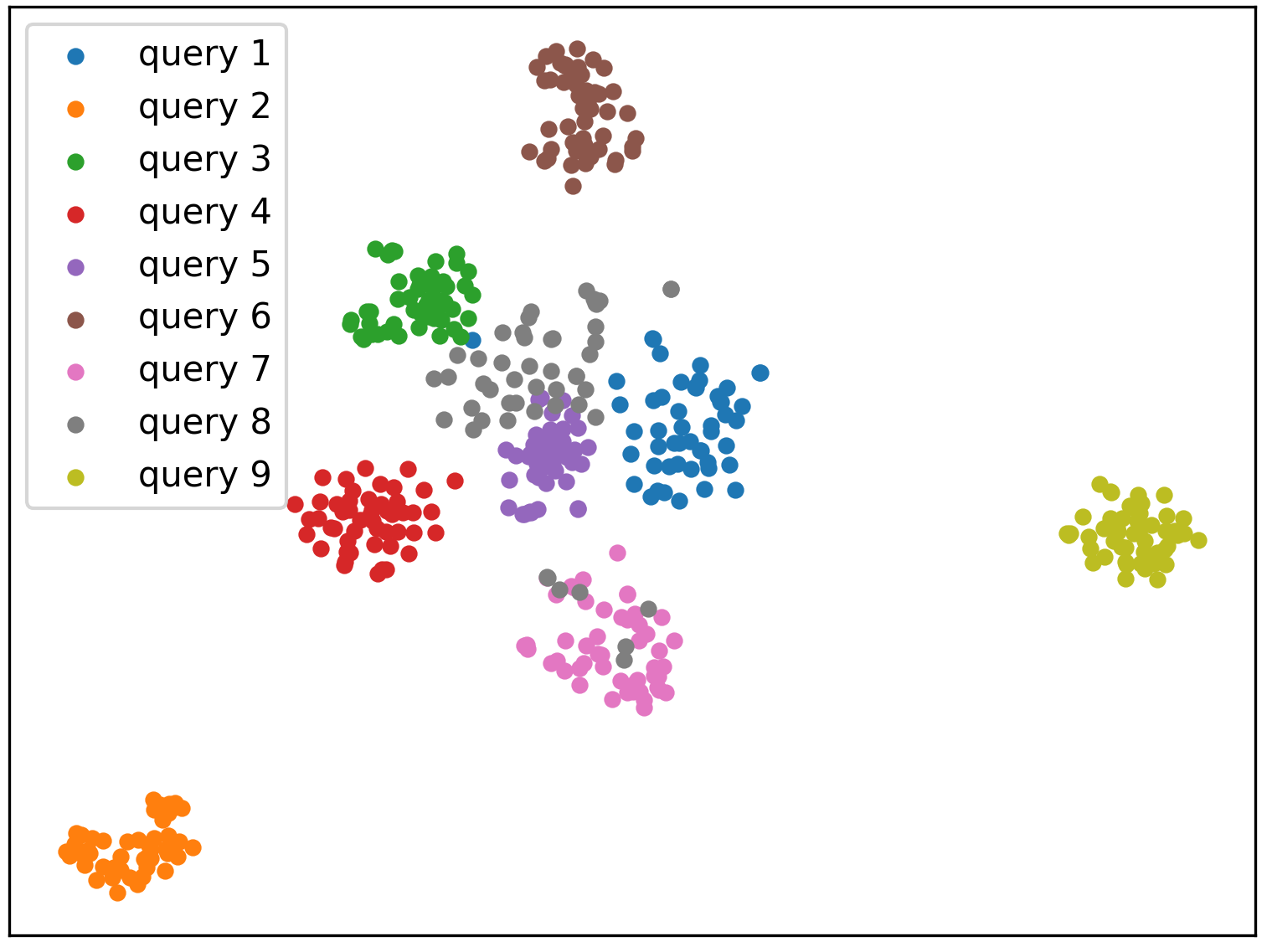}}\subfigure[DaBR (epoch=100)]{
\includegraphics[width=0.32\linewidth]{fig/ours_100.png}}
\caption{Visualization of the embeddings of tail entities using t-SNE. 
A point represents a tail entity. 
Points in the same color represent tail entities that have the same $(h_r, r_j )$ context.
}
\label{Fig:mutliembedding}
\end{figure*}

Figure \ref{Fig:mutliembedding} shows more visualization results.

\end{document}